\documentclass[journal]{IEEEtran}
\usepackage{cite}
\ifCLASSINFOpdf
   \usepackage[pdftex]{graphicx}
\else
\fi
\usepackage[cmex10]{amsmath}
\usepackage{algorithmic}
\usepackage{array}
\ifCLASSOPTIONcompsoc
	\usepackage[caption=false,font=normalsize,labelfont=sf,textfont=sf]{subfig}
\else
	\usepackage[caption=false,font=footnotesize]{subfig}
\fi
\usepackage{fixltx2e}
\usepackage{rotating}
\usepackage{url}
\usepackage{stfloats}
\usepackage{url}
\usepackage{multirow}
\usepackage{authblk}
\usepackage{footnote}
\usepackage{xcolor}
\usepackage{booktabs}
\newcommand{\ignore}[2]{\hspace{0in}#2}

\begin{document}
%
% paper title
% can use linebreaks \\ within to get better formatting as desired
% Do not put math or special symbols in the title.
\title{Review on Classification Techniques used in Biophysiological Stress Monitoring}
%
%
% author names and IEEE memberships
% note positions of commas and nonbreaking spaces ( ~ ) LaTeX will not break
% a structure at a ~ so this keeps an author's name from being broken across
% two lines.
% use \thanks{} to gain access to the first footnote area
% a separate \thanks must be used for each paragraph as LaTeX2e's \thanks
% was not built to handle multiple paragraphs
%

\author{Talha Iqbal$^{1}$, Adnan Elahi$^{2}$, Atif Shahzad$^{1,3}$, William Wijns$^{1,4}$}% <-this % stops a space
\affil{$^{1}$Lambe Institute of Transitional Research, University of Galway, H91 TK33, Ireland\\$^{2}$Electrical and Electronics Engineering, University of Galway, H91 TK33, Ireland\\$^{3}$Centre for Systems Modelling and Quantitative Biomedicine (SMQB), University of Birmingham, Birmingham, United Kingdom\\$^{4}$CÚRAM Center for Research in Medical Devices, H91 W2TY Galway, Ireland\\$^{*}$Corresponding Author: Talha Iqbal (t.iqbal1@nuigalway.ie)}
\maketitle

% As a general rule, do not put math, special symbols or citations
% in the abstract or keywords.
\begin{abstract}
Cardiovascular activities are directly related to the response of a body in a stressed condition. Stress, based on its intensity, can be divided in two types i.e. Acute stress (short term stress) and Chronic stress (long term stress). Repeated acute stress and continuous chronic stress may play a vital role in inflammation in circulatory system and thus leads to a heart attack or to a stroke. In this study, we have reviewed commonly used machine learning classification techniques applied to different stress indicating parameters used in stress monitoring devices. These parameters include Photoplethysmograph (PPG), Electrocardiograph (ECG), Electromyograph (EMG), Galvanic Skin Response (GSR), Heart Rate Variation (HRV), skin temperature, respiratory rate, Electroencephalograph (EEG) and salivary cortisol, used in stress monitoring devices. This study also provides discussion on choosing a classifier, which depends upon number of factors other than accuracy, like number of subjects involve in an experiment, type of signals processing and computational limitations.   
\end{abstract}

% Note that keywords are not normally used for peerreview papers.
\begin{IEEEkeywords}
bio-physiological stress, classification techniques, classifiers, stress parameters, heart attack, physical stress, mental stress, stress monitoring and recognition.
\end{IEEEkeywords}

% For peer review papers, you can put extra information on the cover
% page as needed:
% \ifCLASSOPTIONpeerreview
% \begin{center} \bfseries EDICS Category: 3-BBND \end{center}
% \fi
%
% For peerreview papers, this IEEEtran command inserts a page break and
% creates the second title. It will be ignored for other modes.
\IEEEpeerreviewmaketitle

% The very first letter is a 2 line initial drop letter followed
% by the rest of the first word in caps.
% 
% form to use if the first word consists of a single letter:
% \IEEEPARstart{A}{demo} file is ....
% 
% form to use if you need the single drop letter followed by
% normal text (unknown if ever used by IEEE):
% \IEEEPARstart{A}{}demo file is ....
% 
% Some journals put the first two words in caps:
% \IEEEPARstart{T}{his demo} file is ....
% 
% Here we have the typical use of a "T" for an initial drop letter
% and "HIS" in caps to complete the first word.
%\IEEEPARstart{T}{his} demo file is intended to serve as a ``starter file''
%for IEEE journal papers produced under \LaTeX\ using
%IEEEtran.cls version 1.8 and later.
%% You must have at least 2 lines in the paragraph with the drop letter
%% (should never be an issue)
%I wish you the best of success.
%
%\hfill mds
% 
%\hfill December 27, 2012
\section{Introduction}
Stress can be of two types, Physical Stress and Mental Stress. Physical stress is often caused by a poor diet, sleep deprivation, too much work or may be due to illness. Mental stress is triggered due to worrying about illness of a loved one, death of closed ones, retirement or money or being fired from work \cite{I1}. Generally, most of the stress comes from our daily responsibilities. Work pressure and obligations, which are mental and physical, are not always noticeable to us. In response to stress, incurred in daily life, our body automatically alters our blood pressure, respiration, heart rate, blood flow to muscles and our metabolism. The response tries to help our body to react fast, yet efficiently, to a high pressure situation \cite{I2}.

Stress situation can become threat to our well-being and health, if no adjustments are made in time to cope with its effects. It is very important to realize that all the external events. It does not matter how we are perceiving those events. These events can cause stress and may cause you to feel 'being out of control'.

%\subsection{Effect of Stress on Our Body}
Fatigue, headaches, sleeping problems, digestive problems, and muscle tension are some most common effects of stress. A long term and un-managed stress may cause heart disease, high blood pressure, diabetes, and obesity \cite{sb}. Stress may also cause anxiety, restlessness and inability to focus. These mental and physical changes may effect our weight loss progress as eating junk food can be used to cope stress situation \cite{I3}. Several people gets addicted to tobacco or illegal drugs as they use it as a stress management mechanism. If stress state period goes for very long time, it does increases the chances of having heart attack, hypertension or stroke in a person \cite{I4}.

According to American Psychological Association (APA) stress is linked to 6 leading causes of deaths including heart disease, depression, anxiety disorder, diabetes etc. Center for Disease Control and Prevention reported that 110 million people die every year as a direct result of stress i.e. 7 people in every 2 seconds.

%\subsection{Importance of Stress Monitoring}
We need to realize that without a proper monitoring and management of stress, the situation will get more difficult to contain. People are getting depressed, are very easily angered, and they have started to withdraw to themselves. As stress is becoming one of the movers, performance of people has declined. People thinks this economic recession will stay like this and there is no way to fight it. All these situations can be avoided if people knew how they can fight and win from the effects of stress.

Recognition of high stress state is very essential. For this purpose, one can monitor stress using physiological indicators of stress such as increasing heart rate, blood pressure, respiratory rate, sweaty hands and fast pulse rate. Unfortunately, most people, cannot recognize these physical symptoms. Such people can use stress monitoring devices that will inform them in-time about increase in their stress level and thus they will be able to control it before hand by either doing meditation or by exercise. Literature suggests that following parameters, individually or with different combinations, can be considered for stress monitoring and are discussed below:  

\begin{itemize}
	\item \textbf{PPG:} A Photoplethysmograph (PPG) is light based plethysmogram which is used to detect changes in volume of the blood running in micro-vascular system of the body. Usually, a pulse oximeter is used to obtain PPG. A pulse oximeter illuminates skin using light emitting diode (LED) and calculates change in absorption of light through photo-diode. As flow of blood to the skin is modulated with different types of other physiological systems, thus, we can use PPG to monitor hypovolemia, breathing and conditions of other circulatory system \cite{A9}. For monitoring of the stress, PPG signal obtained from pulse oximeter is used to calculate the oxygen level with-in the body of the human, often called oxygen saturation. For normal person this oxygen saturation level is in-between 95 to 100\%. Saturation level below 90\% is considered abnormal and can lead to a clinical emergency \cite{O1}. The waveform of PPG signal differs patient to patient and to the location as well as the manner in which pulse oximeter is attached. 
	\item \textbf{ECG:} Electrocardiography (ECG, also known as EKG) is a method used to record the electrical activities of the heart with respect to time by placing electrodes on the skin. These electrodes detects small electrical changes under the skin that are generated by depolarization and re-polarization of heart muscles in electrophysiologic pattern for each heartbeat. ECG is commonly used to detect cardiac problems. In stress monitoring application, an ECG signal is used to calculate Heart Rate Variation(HRV). At stress state, the heart rate varies significantly thus can be monitored by using HRV. 
	\item \textbf{EMG:} Elecromyography (EMG) is a technique for recording and evaluation of the electrical activities produced by muscles of skeleton. Instrument called electromyograph is used to generate a record called electromygram. EMG is the recording of the muscle cells whenever these cells are activated electrically or neurologically. Medical abnormalities such as Polymyositis and Muscular dystrophy, recruitment order, and activation levels can be analyzed using EMG signals. Moreover, we can also analyze biomechanics of animal or human movement.
	\item \textbf{GSR:} Galvanic Skin Response (GSR) is used to measure skin's electrical conductance. Sympathic nervous system can be triggered through strong emotion, which results in more sweat being released by sweat glands. The signal is obtain using two electrodes attached to two fingers of same hand and often used to monitor the quality of sleep.
	\item \textbf{EEG:} Electrophysiological method for monitoring and recording electrical activities of the brain is called electroencephalography (EEG). It is, most of the time, non-invasive method with electrodes located on the scalp but sometimes we can use invasive electrodes. Basically, voltage fluctuations are measured through EEG that causes ionic current in the neurons of brain. These readings are analysed with respect to a period of time and diagnosis are made on spectral content or event-related potentials.
	\item \textbf{Respiratory Rate:} Respiration rate is defined as number of times someone takes breath in 60 seconds, mostly calculated by observing the rise and fall of the subject's chest. Measuring the respiration rate while person is sleeping is the most difficult task and cannot be done using general lab devices. Respiratory rate can also be measured from blood volume pulses. There are two ways to do so; by calculating time change in-between two successive heart beats and change in the amplitude of the blood volume. The infrared sensors on the ring takes samples at 250 Hz from arteries and capillaries of finger and thus does not disturbs sleeping subject \cite{RR}. The collected samples shows inter-beat-interval, yielding data on heart rate variability, heart rate and respiratory rate.
	\item \textbf{Salivary Cortisol:} Cortisol is a steroid hormone, which is produced in adrenal gland by zona fasciculata in response to stress. For a normal person, concentration of salivary cortisol ranges from 10.2 to 27.3 with $+/-$ 0.8 $nmol/L$ in morning and ranges from 2.2 to 4.1 with $+/-$ 0.2 $nmol/L$ in night \cite{SC1}. High level of cortisol circulating the human body shows sustained stressed condition of the human and may create allostatic load. This allostatic load can cause various physical changes in the body. Changed level of cortisol can be observed in form of mood disorders, anxiety disorder, illness, stomach pain, fear and other physiological and psychological disorders \cite{sc2}. Collection protocols and approved collection method of salivary cortisol are defined by Salimetrics USA \cite{SU}.
\end{itemize}

Stress can be triggered by using a questionnaire or can be caused due to physio- and sociological factors. Different classification algorithms are used to recognize stress state. In this study, we will focus more on different machine learning techniques as these algorithms are more accurate and popular as well as state-of-the-art way of monitoring and recognizing the stress in humans.

\section{Search Methodology}
Studies only using machine learning algorithms for classification of mental and physical stress with an accuracy of more than 50\% were selected. All the search was done using Google Scholar, Pub Med, IEEE digital library and Cochrane library. The search terms used are as follow:

\begin{itemize}
	\item Stress monitoring using medical devices
	\item "stress" monitoring wearable systems
	\item medical devices for "stress" monitoring
	\item devices to measure mental and physical stress
	\item Different techniques to measure physiological and mental stress and
	\item stress detection health devices
\end{itemize}

Search database yielded 105 studies, with no duplicate study. Studies published between 2000 to March 2019 were taken into the account for literature search. Selection of literature for inclusion in this paper was divided into three steps.
\begin{itemize}
	\item In first step, studies related to recognition of mental and physical stress using machine learning algorithms were included by reviewing paper's title and abstract.
	\item In the second step, studies that did prediction of stress using machine learning techniques, directly, were selected.
	\item In the final step, a deep study was done to acquire information about input data, methods of measurements, classification methodology, tools and number of subjects used for experimentation, accuracy achieved and conclusion drawn by the authors of the papers.
\end{itemize}

Most frequently used machine learning algorithms in these 105 papers are selected for discussion, which are 21 different techniques. For result comparison, in first stage, 23 studies were removed on the bases of information in their title and abstract. In second stage, after thorough studies, 58 studies were excluded due to lack of any or multiple information required, described in previous paragraph. Finally, only 24 studies met the selection criteria and are included in this review.

%\section{Stress Monitoring and Recognition Algorithms}
% needed in second column of first page if using \IEEEpubid
%\IEEEpubidadjcol 
\section{Questionnaire Methods}
A questionnaire method is mostly used to measure mental stress. In this method, stress is triggered in the subject using some questions and recording their response time along with PPG and EEG signals. These response signals are then fed to a neural network to classify the stress states into low, normal and highly stressed states. Such assessment of the stress is often called subjective assessment. Most of the research community uses a "Stroop test" to measure the mental health of the subject. Stroop test is basically a colour naming activity, mostly designed on computer as a game. In this game, subjects are ask to call name of the colour irrespective of what is written with that colour. For example, word BLUE is written with PURPLE colour, so subject have to say PURPLE as an answer to this question. Stroop test designed by Nagananda et. al. in \cite{R23} uses five colours; BLUE, YELLOW, GREEN, RED and PURPLE and classifies stress into low, medium and high stress levels using simple neural network.

Beside stroop test, many researches designs their own test questionnaires. As Kallus et. al. \cite{R24} designed a RESTQ that measured the frequency of stress and activities related to stress recovery. The authors designed five different versions of RESTQ based on the types of subjects one want to use. RESTQ-Basic for general usage and had seven stress scales. RESTQ-Sport for athletes with five recovery scales. RESTQ-Coach for coaches, RESTQ-CA for adolescents and RESTQ-Work for subject's work context. Every version had its own time frame of three days/nights or seven days and nights. The output is indicated on the scale of 0 to 6 i.e. never to always, respectively. 

Boynton et. al. \cite{R25} presented a very interesting study about selection, design and development of a self defined questionnaire. The authors argued that anyone can design a list of questions and print it but designing a well effective and generalized questionnaire needs creative imaginations and careful planning. Authors also discussed different perspectives of a questionnaire that should be taken into account while designing or developing it. A new questionnaire often fails to provide a high quality generalized data, thus, whenever possible one should use previously validated questionnaires and rephrase them appropriately for their targeted audiences and information they require. Authors concludes that a nicely explained and carefully designed questionnaire will always lead to improved response rates.
 
\section{Machine learning Algorithms}
Recently, development of different machine learning algorithms has greatly helped to develop tools that assists doctors to support patient care and predict any mental disorders. Machine learning techniques are widely used as a decision boundary making tool in complex data analysis of health. Supervised machine learning algorithms produces general hypotheses from externally supplied labelled feature. This hypotheses is then used to make predictions about new incoming features \cite{ml}. Literature included in this review are measured on following criteria:

\begin{enumerate}
\item Should be frequently used classification methods.
\item Should have achieved good classification accuracy ($>$50\%).
\item Should be about classification of physical and mental stress states. 
\end{enumerate} 

Selection of learning algorithm is a critical step. Usually, an algorithm is evaluated on the basis of its number of correctly predicted outcomes over total number of prediction attempts (which is called prediction accuracy). There are three ways to testing a classifier. One is by splitting the training set into training and evaluation set. The ratio of split should be 70\% and 30\%, respectively. Cross validation is the second option to test the performance of the classifier. Here training set is split in mutually exclusive and equal-sized subsets. Classifier is then, for every subset, trained on the union of all subsets. Error rate of every subset is calculated and average error rate determines the classifier performance. There is a special type of cross validation called Leave-one-out. Computationally this method of cross validation is expensive but is used whenever we require greater accuracy in terms of error rate of a classifier. Error rate does depends upon numbers of parameters like size of training set, dimension of problem, hyper-parameters tuning and use of relative features of the problem. Third and final way to measure the performance of the classifier is statistical comparison of classifier's accuracies when trained on specific datasets \cite{A1}.

Supervised machine learning algorithms can be divided into four major types; Logic-based Algorithms, Perceptron-based Methods, Statistical learning Techniques and Support Vector Machines (SVM). Each type of algorithm have different sub-learning algorithms. Some of these learning algorithms are illustrated as follows:

\subsection{Decision Tree Classifier} 
Decision tree does classification by sorting input instances on the basis of feature values \cite{A2}. Each node of the decision tree shows a classified feature from an input instance while each branch shows an assumed nodal value. Classification of instances starting from root and is sorted depending upon their feature values. The best divisor of input training data becomes root node of the decision tree. %Decision tree is said to be over-fitted, when for any learned hypothesis $h$, there exists another $h^{'}$ hypothesis that have greater error rate then learnt hypothesis $h$ whenever it is tested on training dataset but have less error rate when tested on complete dataset.

\begin{figure}[!h]
	\centering
	\includegraphics[width=3in, height = 8in, keepaspectratio]{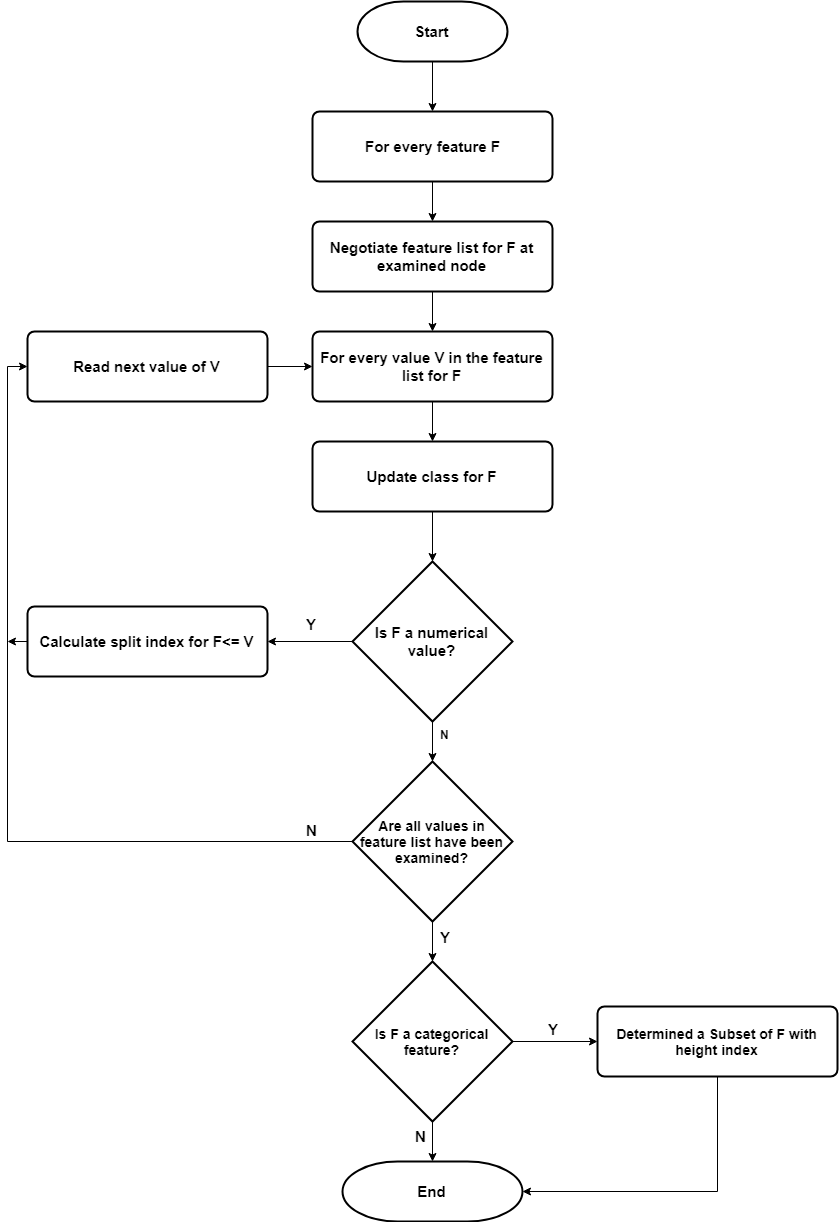}
	\caption{Flowchart of C.45 decision tree algorithm}
	\label{fig_1}
\end{figure}

Fig. \ref{fig_1} shows steps involved in classification using a decision tree. This divide and conquer strategy is efficiently fast and can be an important classifier if we have hundreds of thousands of input instances. \ignore{The only time consuming parameter in this algorithm is sorting aspect of the instances on the basis of numeric features so that we can have an optimal threshold value $t$.} A pseduo-code for designing a decision tree is presented in \cite{A3}. \ignore{One of the critical characteristic of a decision tree is its comprehensibility i.e. allocation of an instance to a specific class can easily be understood by everyone whenever decision tree is used as classification algorithm. Sum of outcomes in the decision tree is 1 (so, each outcome can be thought of a probability value of the instances belonging to the specified class).}

\subsection{Artificial Neural Network Classifier}
\begin{figure}[!h]
	\centering
	\includegraphics[width=2.5in, height= 5in, keepaspectratio]{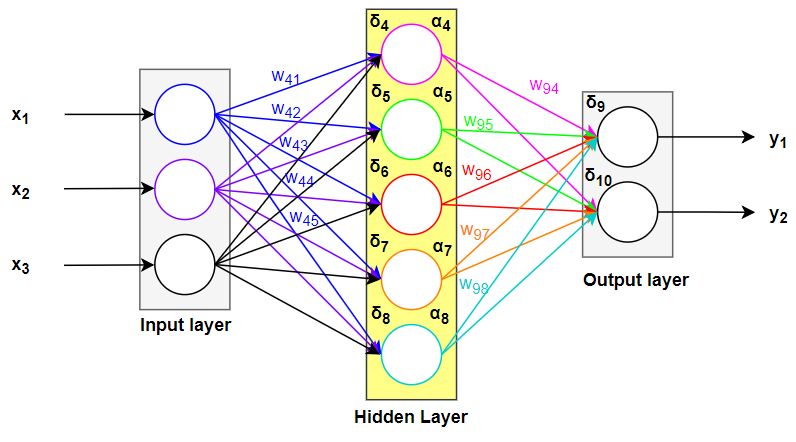}
	\caption{Feed forward ANN}
	\label{fig_2}
\end{figure}

Artificial Neural Network (ANN) are used for classification whenever instances in training dataset can not be linearly separated. An overview of Artificial Neural Network (ANN) is provided in \cite{A4}. ANN is created by connection of many neurons (units) patterned as shown in Fig. \ref{fig_2}.

Neurons of the network are divided into three layers : input layer, which receives incoming information from training dataset;  output layer, that is gives us processed result (most of the time probabilities); and hidden layer, which is in-between input and output layer. If there is one-way communication between neurons of the network i.e. only from input to output, then the network is called feed forward network. Outcome of an ANN depends upon three factors: architecture of the network, weights associated with each neuron in the network, and input along with activation functions used for each neuron. All weights are updated in such a way that it brings the outcome (result of classifier) nearer to the desired output. The most popular algorithm for updating weights is known as Back Propagation (BP) algorithm and is defined as:

\begin{equation} \label{eq1}
\Delta {{W}_{ji}}=\eta \; {{\delta }_{j}} \; {{O}_{i}}
\end{equation}

\ignore{Here: $\eta$ is learning rate, ${O}_{i}$ is output of $i^{th}$ neuron while ${\delta }_{j}$ is output neurons (and/or output of hidden neurons). Each neurons have its own activation value. All the input layer neurons sends their activation value to each neuron in the hidden layer then each neuron of the hidden layer calculates activation value for itself and propagates it to output set of neurons, as shown in Fig. \ref{fig_2}. The overall result of ANN classifier is calculated by summation of all the contributing neurons values, and this contribution is defined by multiplication of weight of the connection with sending and receiving neuron values. ANNs are used in many real-world applications but these networks are less comprehensible i.e. these networks lacks the ability to show the reasons behind specified outcome, effectively.}

\subsection{Bayesian Network Classifier}
\begin{figure}[!h]
	\centering
	\includegraphics[width=3.5 in, height= 5in, keepaspectratio]{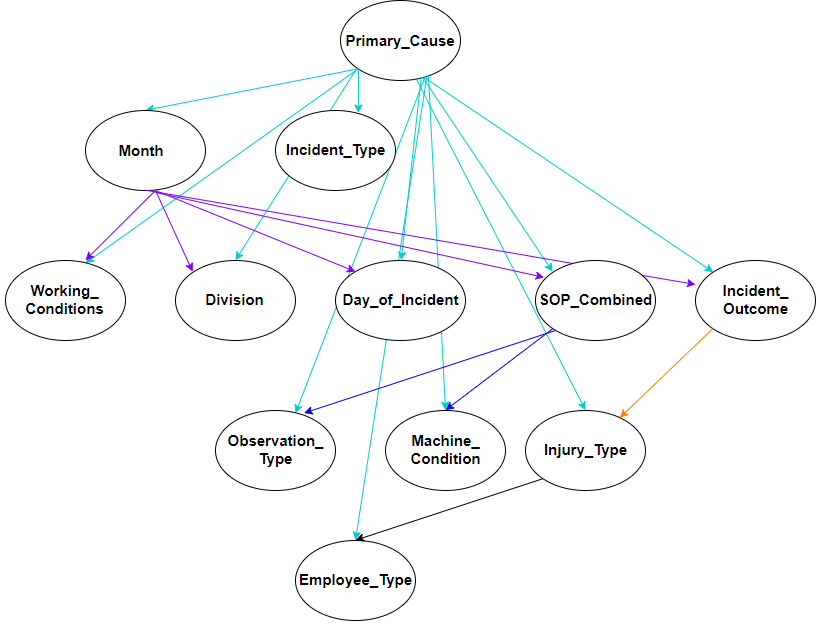}
	\caption{Structure of BN}
	\label{fig_3}
\end{figure}

To represent probability relationships of input instances (features) in form of graphs, Bayesian Network is used. Structure of Bayesian Network (BN) is Directed Acyclic Graph (DAG) and there is one-to-one correspondence between its nodes. Arcs in DAG shows influence of different features on each other. Conditional independences can be detected if there is no arc representing casual influences in-between features or there is no descendants nodes from this node (feature). Typically, learning Bayesian network is two-fold task: first, learn DAG structure and how BN structure is created using DAG, then determine BN parameters. Pseudo-code for training BN is presented in \cite{A3}. Fig. \ref{fig_3} shows structure of general Bayesian network.

\ignore{The advantage of BN over Decision Tree and Artificial Neural Network is that BN takes prior information given about the problem into account and makes structural relationships between its features. There are only two problems with BN classifier; it is not suitable for the datasets having many features, as demonstrated in \cite{A5} and discretization of, almost all, the numerical features.}

\subsection{Naive Bayesian Classifier}
\begin{figure}[!h]
	\centering
	\includegraphics[width=2.2in, height= 4in, keepaspectratio]{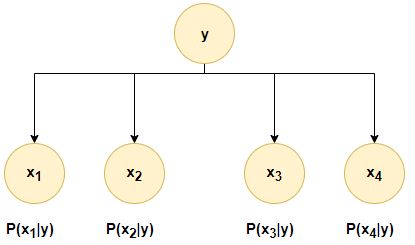}
	\caption{Structure of NB}
	\label{fig_4}
\end{figure}

Naive Bayesian classifier is very simpler form of Bayesian network. Naive Bayesian (NB) has only one parent node in its DAGs, which is an unobserved node, and have many children nodes, representing observed nodes (Fig. \ref{fig_4}). NB works with a strong assumption that all the child nodes are independent from its parent node and thus, one may say that Naive Bayesian classifier is a type of estimator. Mathematically,

\begin{equation} \label{eq2}
R=\frac{P(i|X)}{P(j|X)}=\frac{P(i)P(X|i)}{P(j)P(X|j)}=\frac{P(i)\prod{P({{X}_{r}}|i)}}{P(j)\prod{P({{X}_{r}}|j)}}
\end{equation}

From equation \ref{eq2}, one can conclude that larger probability value will indicate that class label assign to a feature (child node) is its actual label. The threshold for classification is as; if value of $R > 1$ then predict $i$ otherwise predict $j$.

\ignore{This assumption of all the children nodes are independent from its parent node (almost every-time) proves to be wrong and thus, this is why naive Bayesian classifiers are not a much accurate as other sophisticated learning algorithms, like ANN, are. However, a big advantage of naive Bayes classifier is its small computational time for its training.}

\subsection{k-Nearest Neighbour Classifier}
\begin{figure}[!h]
	\centering
	\includegraphics[width=2.5in, height = 5in, keepaspectratio]{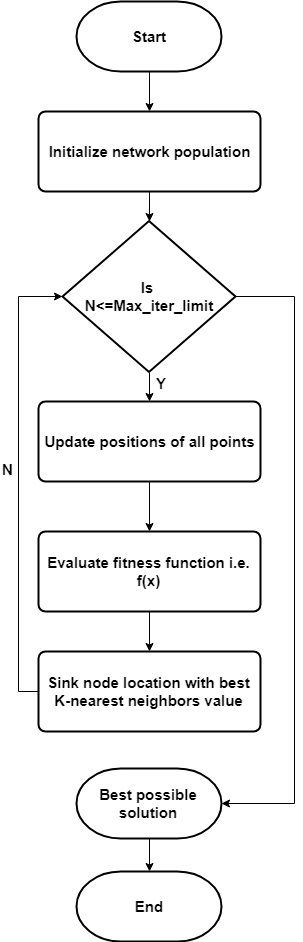}
	\caption{Flowchart of kNN classifier}
	\label{fig_5}
\end{figure}

k-Nearest Neighbour (kNN) is one of the simplest instance based learning algorithm. Working of kNN is as; it classifies all the close proximity instances, in a database, into a single group and then when new instance (feature) comes, the classifier observes the properties of the instance and place it into the closest matched group (nearest neighbour). Fig. \ref{fig_5} shows flowchart for working of kNN classifier. For accurate classification, initializing a value to $k$ is most critical step in kNN classifier.

\ignore{k-Nearest Neighbour might not classify an instance (feature) accurately if the initial value $k$ is specified in-correctly because of the two reasons; if there is noise present in the input feature dataset, it wins majorities of votes while kNN predicts a class of the feature and thus, results in inaccurate classification. To tackle such kind of problem one may choice a larger value of $k$. If we have so small region from classification of an instances that surrounding region wins majority of the votes and that is where we can not use larger value of $k$. A small value of $k$ is used to solve this problem.} 

\ignore{Easiness of implementation and simple structure of kNN are the two major advantages of kNN based classifier while drawback includes high computation cost, especially when we have large dataset and is very sensitive to the noise/irrelevant features.}

\subsection{Nearest Cluster Classifier}
Nearest Cluster Classifier is a new classification technique proposed to reduce the training set of k-Nearest Neighbour (kNN) and enhance its performance by using clustering method, (proposed in \cite{A10}). The main goal of this method is to classify the given test samples according to their nearest neighbour tag.

This algorithm first clusters the given (training) set to a number of different partitions. After getting these partitions, different clustering algorithms are executed to eliminate large number of clusters from those partitions. Then, central label of previously produced clusters is calculated using majority vote method between patterns of class labels in the cluster. A set of most accurate clusters are used as training set of the final 1-NN classifier. In last, the class label of upcoming test sample is calculated according to class label of his nearest cluster centre \cite{ncc}. \ignore{Computationally, NCC algorithm is faster than kNN by K-times. As far as accuracy is concerned, NCC is not so worst than kNN. In-fact, in some cases, NCC technique performance better than kNN classifier, see Table 3 in \cite{A10}.}

\subsection{Learning Vector Quantization Classifier}
The Learning Vector Quantization (LVQ) algorithm is artificial neural network based algorithm which enable us to choose number of training features and learns how those features should look like. LVQ is more likely a collection of codebook vectors. These vectors are consists of a list of numbers which have same input as well as output features as their training set. 

\begin{figure}[!h]
	\centering
	\includegraphics[width=3.5in, height= 5in, keepaspectratio]{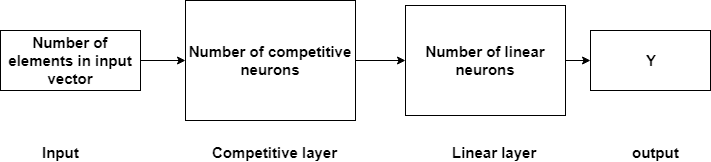}
	\caption{LVQ network architecture}
	\label{fig_16}
\end{figure}

In form of neural network, every vector of the codebook is considered as a neuron, each feature on the codebook vector is considered weight and collection of vectors of codebook makes a network \cite{lvq}. Prediction procedure of LVQ is same as used in kNN (k-Nearest Neighbour) algorithm. For  prediction, all the vectors in the codebook is searched and most similar $K$ is found. Then output is summarized for those selected $K$ instances. By default, the value of $K$ is considered as 1 and best matched vectors in codebook are called Best Matching Unit (BMU). Fig. \ref{fig_16} shows general architecture of LVQ classifier.

\ignore{Similarity between the $K$ instances of the training dataset and new input instance is measured with euclidean distance formula. Mathematically, written as:
	
	\begin{equation} \label{ed}
	d(X,X_{i}) = \sqrt{\sum_{i=1}^{n}({X-X_{i})^2}}
	\end{equation}    
	
Before applying LVQ, the training data is pre-processed in the same way as in kNN algorithm. LVQ can be used for binary classification as well as multiple-class classification. General steps involved in LVQ algorithm are; one should perform multiple passes of complete dataset over the vectors of codebook, Then find multiple best matches by extending LVQ during the learning (training) process, Normalize the input to 0 to 1 scale and select random value between 0 and 1 to initialize the vectors in codebook, and do feature selection carefully, as it will reduce the dimensionality of the input data as well as improve the overall accuracy of the algorithm.}

\subsection{Kohonen Self-Organizing Map Classifier}
The Kohonen Self-Organizing Map (KSOM) learning algorithm is originally presented in 1982, and uses vector quantization with similarities of patterns \cite{A8}. This algorithm is generally used for clustering problems having very complex dataset. Sizes of topology map and learning are the two parameters that should be taken into account before designing a KSOM based classifier. 

\begin{figure}[!h]
	\centering
	\includegraphics[width=3.5in, height= 5in, keepaspectratio]{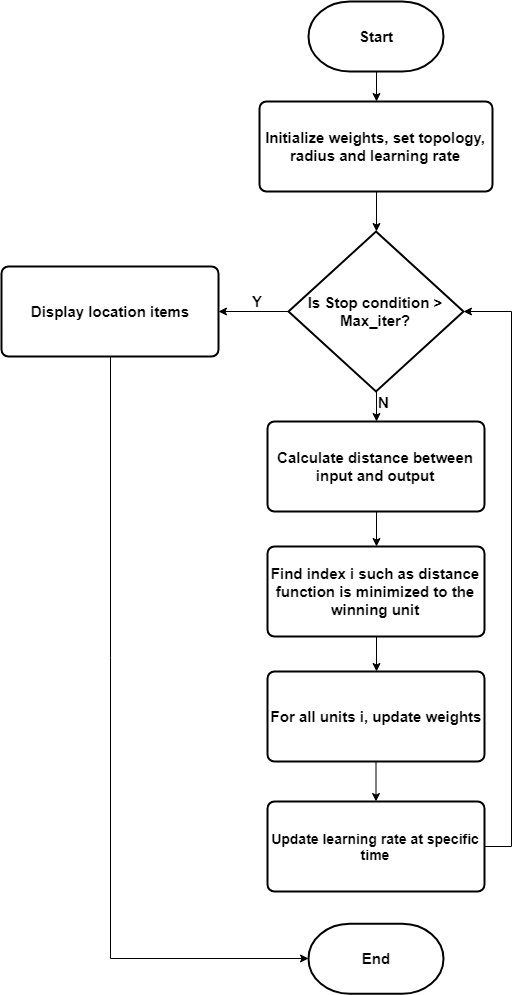}
	\caption{Steps in original KSOM learning algorithm}
	\label{fig_7}
\end{figure}

\begin{figure}[!h]
	\centering
	\includegraphics[width=2.5in, keepaspectratio]{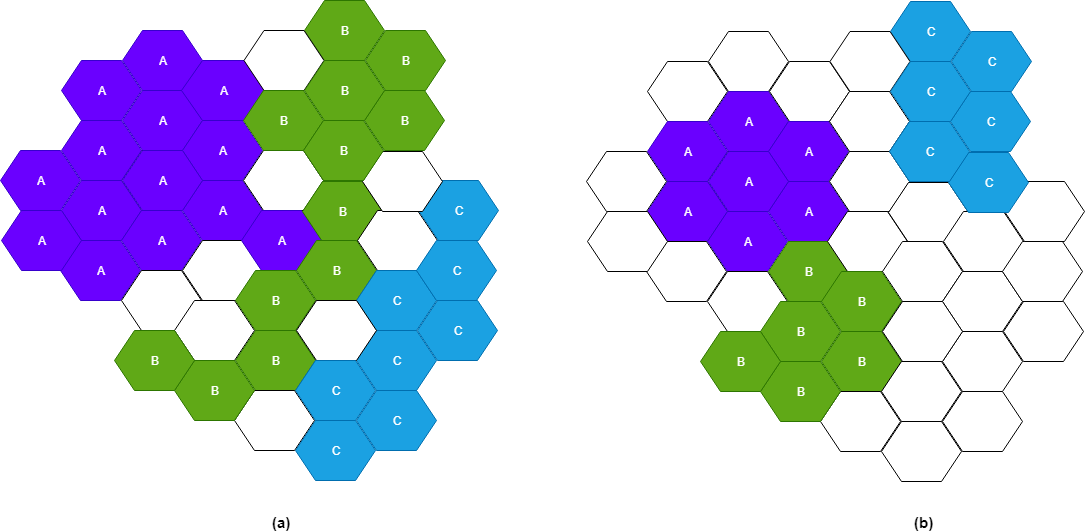}
	\caption{Result of (a) Original KSOM (b) Revised KSOM algorithm on Iris dataset}
	\label{fig_8}
\end{figure}

Steps used for designing KSOM network for classification is shown in Fig. \ref{fig_7}. Complete dataset is repeatedly trained using different sized maps and a suitable size is found for accurate cluster classification. \ignore{As KSOM does clustering on the basis of similarities between instances (features), it can handle a large dimensional dataset. Similar instances are grouped into one node on the map while dissimilarities increases the separating distance between two objects of the map.
	
The algorithms, in start, calculates the distance between all clustering nodes present on map topology. All nodes are arranged in form of lattice, and in input layer, each node is connected to input data.} Euclidean Distance model is used to calculate distance between two nodes. Even grouping and clustering done by KSOM is quite accurate but as every clustered group lies close to another group, sometimes, leads to overlapping of cluster and problem of non-linear separation. To tackle this problem, \cite{A9} presents a variant of original KSOM classification algorithm. They used different approach for distance calculation and Fig. \ref{fig_8} shows result comparison of original and revised KSOM classifier.

\subsection{Principal Component Analysis}
Principle Component Analysis (PCA) is generally used to reduce dimensions of high-dimension dataset to a small subspace before using it to train any learning or classification algorithms. PCA transforms data into a new co-ordinate system having low dimension subspace. In this co-ordinate system, first axis represents first principle component that represent greatest aggregate of variance in a dataset \cite{pca}.

\ignore{Let us suppose, we have a dataset with two variables $x_{1}$ and $x_{2}$ and we want to find first principle component which shows highest aggregate of variance. Easiest way to do that is to draw a straight line (oval lengthwise) and project the reflection of each component (feature) on the line. So, in this way, our two dimensional data is converted to one dimensioned data. This single line is our first principle component having most important information (variance) for both the dataset variables i.e. $x_{1}$ and $x_{2}$. We should notice that if we have only one principle component then majority of the data will be destroyed. So, it is important to know how many principle components are required to cover all the data (if we do not want to loss any information at all).}

\begin{figure}[!h]
	\centering
	\includegraphics[width=3in, keepaspectratio]{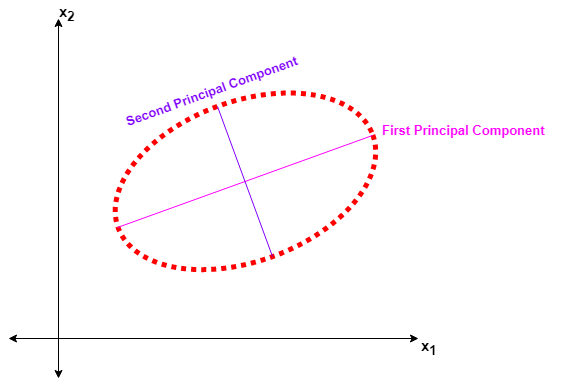}
	\caption{Two Principle Components of dataset having two variables $x_{1}$ and $x_{2}$}
	\label{fig_9}
\end{figure}

\ignore{If we want to capture best variance, the second principle component should always be kept orthogonal to its first principle component.} From fig. \ref{fig_9}, we can see that by calculating two principle components, we can cover the variance of whole dataset and all captured elements are independent of each other. \ignore{Before PCA, it is better to normalize the whole dataset first because PCA seeks highest variance and if data is not properly normalized, attributes having largest value will dominate overall variance (first component). By normalization, almost all the attributes gets an opportunity to play role in principle component analysis.}

\subsection{Linear Discriminant Analysis}
Linear Discriminant Analysis (LDA) is most commonly used dimensionality reduction algorithm. LDA is used during pre-processing step in a pattern classification and other machine learning applications. \ignore{ Training of an algorithm on high dimensional dataset often leads to over-fitting of the algorithm. LDA projects given dataset onto a low dimensional space having good class separability to save the algorithm from over-fitting and also reduces computational cost.}

\begin{figure}[!h]
	\centering
	\includegraphics[width=2.5in, keepaspectratio]{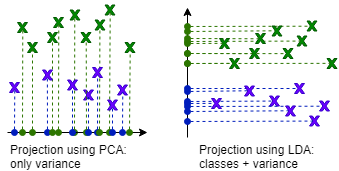}
	\caption{Comparison of PCA and LDA}
	\label{fig_14}
\end{figure}

LDA is calculated in five steps \cite{lda}; compute d-dimensional mean vectors for each class in a given dataset, compute the in-between-class scatter matrix and within-class scatter matrix, then compute Eigenvectors and their corresponding eigenvalues for both scatter matrices, select linear discriminants for the new feature subspace by sorting the eigenvectors in descending order using eigenvalues, and on final step, transforming the samples onto new subspace by simply doing matrix multiplication. Generally, LDA \ignore{is almost same as PCA. The only difference is in addition to calculation of component axis having maximum variance between input data (as in PCA), we have to} calculate an extra axis that shows maximum separation between the two classes, as shown in fig. \ref{fig_14}. \ignore{Dimensionality reduction not only reduces computational cost of the algorithm but also help to avoid over-fitting through minimizing the error in estimation of parameters.}

\subsection{Logistic Regression}
Logistic Regression is one of the simplest machine learning algorithm mostly used for binary classification problem. This algorithm can be implemented easily and is used a baseline algorithm for other two-class classifiers. There are three types of logistic regression algorithms \cite{lr}:

\begin{enumerate}
	\item If the targeted features has only two outcomes like Spam or Not spam emails or Diabetic or Not Diabatic, then this problem is solved by using binary logistic regression.
	\item If the targeted features has three or more than three nominal categories, then multinominal logistic regression. For example, prediction of type of clothing.
	\item If the targeted features has three or more than three ordinal categories, then ordinal logistic regression like rating any product between 1 to 5.
\end{enumerate}

\begin{figure}[!h]
	\centering
	\includegraphics[width=2.5in, keepaspectratio]{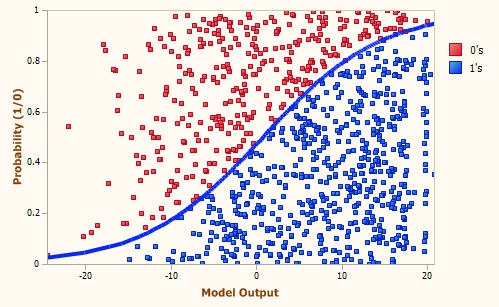}
	\caption{Classification using Logistic Regression \cite{flr}.}
	\label{fig_10}
\end{figure}

Logistic regression estimates and describes relationship between independent and dependent binary features within a dataset. \ignore{This algorithm can be considered as a special case of linear regression in which target (output) feature is categorical. Probabilities of the occurrence of binary inputs (features) are predicted with the help of logit function. Logistic regression method have two properties; dependent features in logistic regression always follows Bernoulli Distribution and maximum likelihood is used for estimation and prediction.} Fig. \ref{fig_10} shows classification boundary calculated by logistic regression algorithm. 

\ignore{Logistic regression can not handle a dataset with large number of categorical variables (features) and tends towards over-fitting. The algorithm also require transformation of non-linear features, when we have non-linear problem to solve using algorithm. Advantage of using logistic regression includes; low computational time and easy to implement. Moreover, this algorithm do not require normalization of features and provides probability scores for each observation in the dataset.}

\subsection{ZeroR and OneR classifier}
ZeroR is the simplest classification technique, also known as Zero Rule algorithm, that relies only on target and ignores all predictors. This algorithm only predicts majority class of the given dataset. It is used for determining a baseline performance as a benchmark for other classification methods, although there is no predictability power in ZeroR. This algorithm construct a frequency table for the target and select its most frequent value \cite{zr}.

OneR is abbreviation of One Rule. It is simple yet accurate classification technique which can produce one rule for every predictor present in the data. Then rule with the smallest error rate is selected and named 'one rule'. Rule for the predictor is created by construction of frequency table having two columns; target and its frequency. OneR has slightly less accuracy as compare to state of the art algorithms used for classification but is too easy for humans to interpret. Following is the pseudo-code for designing a oneR algorithm \cite{or}:

\begin{enumerate}
		\item {For every predictor,}
		\item {For every value of predictor, make following rule;}
		\begin{itemize}
			\item {Count appearance of each target value.}
			\item {Find most frequently appeared class.}
			\item {Make the rule assignment of the selected class to the predictor value.}
			\item {Calculate error rate for each predictor.}
		\end{itemize}
		\item {Choose the predictor with smallest error rate.}   
\end{enumerate}

\subsection{Multi-Layer Perceptron Classifier}
Multilayer perceptron are build using number of neurons connected in different layers, mostly to solve a non-linear classification problem. Each perceptron is used to categorize small linearly discrete regions of input problem while output of each perceptrons is integrated together to generate final output.\ignore{ If a hard limiting like step function is used as an activation function of the neurons to produce output, the information of the real input is changed as it moves towards inner neurons. As a solution to this problem, a continuous function like sigmoid is used.}

\begin{figure}[!h]
	\centering
	\includegraphics[width=3.5in, keepaspectratio]{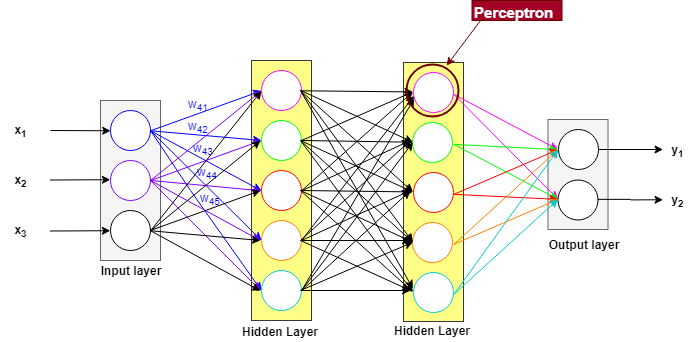}
	\caption{Structure of MLP Classifier}
	\label{fig_13}
\end{figure}

Neurons in multilayer perceptrons are organised as input layer, one or more hidden layers and the output layer, as shown in fig. \ref{fig_13}. The learning rule for this technique is called back-propagation rule or generalised delta rule. \ignore{According to this rule, algorithm repeatedly calculates error function for every input and propagates this error back to the previous layer. The weights of any particular node is tuned directly in accordance to error value of the units connected to it. A single layered perceptron draws only a single boundary between classes.}

\ignore{Advantages of MLP classifier includes; generalization, fault tolerance, have  high accuracy in applications like speech synthesis, and pattern recognition but MLPs are computationally very expensive. Also, there is no guaranteed solution for every problem and have scaling problem i.e. can not scale it up from small research system to any larger real system.}

\subsection{Genetic Algorithm}
\begin{figure}[!h]
	\centering
	\includegraphics[width=2.5in, height= 5in, keepaspectratio]{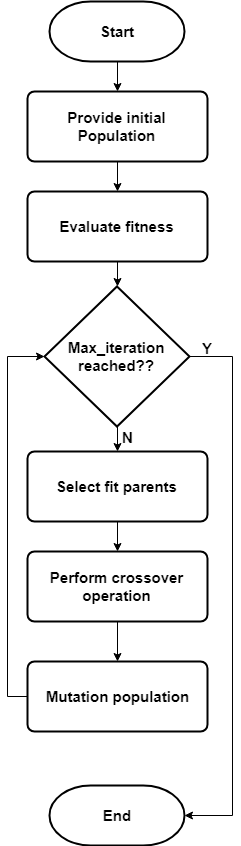}
	\caption{Steps involved in Genetic Algorithm (GA)}
	\label{fig_15}
\end{figure}

Genetic Algorithm (GA) is an optimization technique that follows principles of natural selection and genetics. This technique is frequently used for finding an optimal or nearer optimal solution of complex problems that otherwise may take a long time to get solved. Genetic algorithms basically have pool of possible solutions to a given problem. These solution undergoes mutation and recombination just like in natural genetics, making new children, and this process is recurred again and again over different generations. Every individual or candidate solution is provided with a fitness value (label) depending upon their objective function. The fittest individual is given more chance to mate and produce more fitter offspring (individual). Fig. \ref{fig_15} shows general steps involved in Genetic Algorithm.

\ignore{Genetic algorithms are random in nature but they do perform better than any local random search that just tries different solutions and keeps track of best one, for a long time, which consumes a lot of memory. Main motivations of using GAs includes their ability of solving complex and difficult problems. For example, NP-Hard problems, if you have failing Gradient Decent Method like stuck at local optimal point, and if you want to get good solution quickly. GAs do not require any derivative information (mostly un-available for real-world problems), are more efficient and faster than other traditional methods, can optimize continuous as well as discrete functions. Also, can be used to solve multi-objective problems, and can provide different optimal solutions instead of one single solution to a problem.
	
Along with above advantages of GAs, there are also some limitations. For example, these algorithm is computationally expensive as fitness value is calculated repeatedly and quality of the solution provided is not guaranteed to be the best. Moreover, if these algorithms are not implemented properly (initialization of hyper-parameters not done carefully), these algorithm might not even given any optimal solution.}  

\subsection{Decision Forest}
Decision Forest classification technique is a supervised learning technique. This classifier performs best if we have to predict only two target classes at the output. Decision forest is typically implemented using multiple decision tress model. Hyper-parameters of each decision tree model are tuned before training and testing of decision forest classifier. When combined, each decision tree votes for the popular output class, which is mostly used whenever we have ensemble model.

Working of decision forest is as \cite{df}; it creates many individual tress using whole dataset, keeps starting point of each tree different (usually this selection is random), every tree in decision forest gives a non-normalized histogram of frequency labels at the output, these histograms are then summed up and normalized to get probability of each label, and at last, the tree having highest confidence value is given higher weight in the last decision of the forest.

\ignore{Before selecting decision forest as your classifier you should know few hyper-parameters that needs to be initialized before-hand. These parameters includes \cite{df} resampling method; You should know how you want to create each individual tree (could either be $begging$ or $replicate$ technique), specify how to train your model; either you want to train your model using $single$ parameter or you want to train it via parameter $range$, number of decision trees in a forest; increasing the number of trees may increase accuracy but training time will increase, maximum depth of the forest; increasing the depth of the forest may increase precision but may result in over-fitting and also, increases training time, number of splits from each node; determine the number of features in each node, minimum number of samples per leaf; show how many number of cases you require to make a terminal node i.e. leaf of the tree. If you increase this value you ultimately increase the threshold for new rule creation, and know the labels of each instance in the dataset and attach class labels with the whole dataset.
	
Decision forest can creature non-linear boundaries for the classification, can handle data with varied distribution and less effected by the noise in the features. This algorithm is also computationally efficient and uses less memory but can over-fit on given dataset as well as lacks generalization.}

\subsection{Decision Jungle}
Decision Jungles are extension of decision forest. A jungle is consists of ensemble decision Directed Acyclic Graph (DAG). Hyper-parameters to be initialized are \cite{dj} resampling method; you should know how you want to create each individual tree (could either be $begging$ or $replicate$ technique), specify how to train your model; either you want to train your model using $single$ parameter or you want to train it via parameter $range$, number of decision DAGs; define maximum number of graphs which will be created, maximum depth of the decision DAGs; determine maximum depth allowed for each graphs, maximum width of the decision DAGs; determine maximum width allowed for each graphs, maximum number of iteration per decision DAGs; determine number of iterations over given dataset to build each DAG. 

Decision jungle allows branches of the trees to merge thus this algorithm can be generalized. Moreover, this algorithm can create non-linear boundaries during classification and is robust to noisy feature.

\subsection{Random Forest}
\begin{figure}[!h]
	\centering
	\includegraphics[width=3.5in, height=2in]{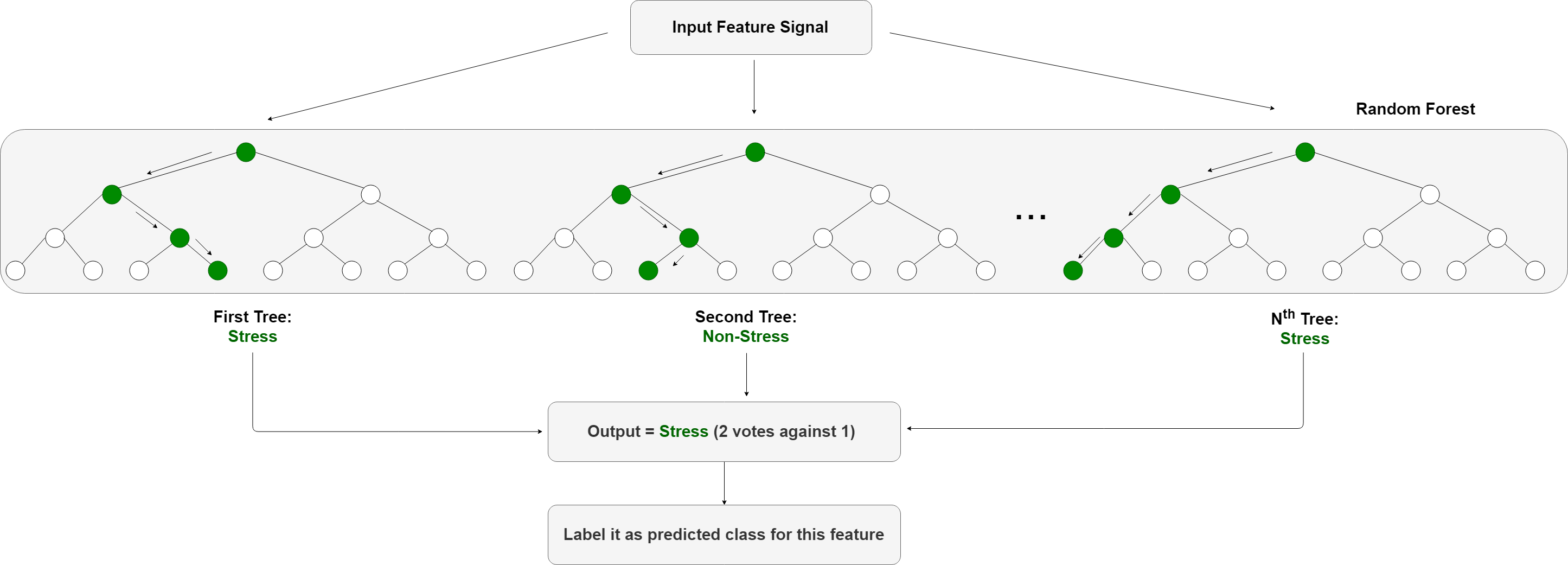}
	\caption{Simplified Random Forest Classification, classifying stress and non-stress}
	\label{fig_17}
\end{figure}

Random Forest is a supervised machine learning algorithm. This algorithm creates a random trees (forest) that are somewhat similar to decision trees and training method selected is always $begging$, as in $begging$ we combine learning models linearly to increase the overall accuracy. While growing new trees, random forest adds more randomness to the existing model. Instead of finding most important target feature for node splitting, this algorithm searches for best feature in the random subset of target features. In this way, we get wide diversity which in-return results in better model. So, as random forest only consider a random subset of features for splitting a node, we can make the trees of the model more random by using random thresholding of every feature rather than looking for the best threshold value \cite{rf}. \ignore{Random forest is very easy yet a handy algorithm as one can use it without tuning the hyper-parameters at the start and default values of hyper-parameters produces accurate prediction (classification). Fig. \ref{fig_17} shows classification of an input instance (feature signal) using random forest classifier.}

\ignore{One of the biggest problem with machine learning algorithms is over-fitting, but random forest classifier, most of time, gets away from this problem because there are sufficient trees in the forest that do not let the classifier to get over-fit. Main limitation of this classifier is slow computation. A large number of trees in the forest makes prediction process really slow, sometime as much slow that it become inefficient to be used for real-time prediction. One have to make trade-off between accuracy and prediction time, as these algorithms trains quickly. Greater accuracy requires larger number of trees in the forest but slowers the prediction, once trained.}

\subsection{One vs All Multiclass Model}
This model is used for prediction of three or more than three classes, specifically when the target outcome is dependant on categorical or continuous prediction variables. This algorithm can also be used for binary classification that needs multiple classes at the output \cite{oam}. Model is created using binary classifiers for each of the multiple class output. Every binary classifier is assigned to individual classes as an complement to all other classes present in the model. Final prediction is done by running all the binary classifiers and selecting the prediction having highest probability score. In essence, a group of individual model is made and results are then merged at the end, to create single model which can predict all classes. Therefore, we can choose any binary classifier as basic classifier for one-vs-all model. To configure such model one should know that there are no configurable parameters of this model. So, optimization should be done in the binary classifier which is to be provided at the input for building this model.

\ignore{The biggest limitation of one-vs-all multi-class model is class imbalance. To understand this problem, let us suppose we have a two class (binary) classification problem. The two classes are named as class A and class B. Suppose dataset, which is to be classified, have 95\% of data that belongs to class A while 5\% of the data that belongs to class B. Now, consider we have trained our model on this data. The classifier predicts that 100\% of the data belongs to class A and will still achieve 95\% of accuracy. In other words, classifier need a more balanced distributed data rather than skewed distributed dataset \cite{oam}.}

\subsection{Ada-boost}
Boosting refers to group of techniques that creates a strong classifier using numbers of weak classifiers. To find a weak classifier, different machine learning based algorithm having varied distribution are used. Each learning algorithm generates a new weak classification rule. This process is iterated many times and at the end a boosting algorithm is formed by combining all newly generated weak classifiers rules to make a strong rule for prediction. There are few steps that should be followed for the selection of right distribution \cite{ab}:
\begin{itemize}
		\item Step 1: Give all the distributions to base learner and assign equal weights to every observation.
		\item Step 2: If first base learner gives any prediction error, then pay more attention to observations causing this prediction error. Then, apply new base learner.
		\item Step 3: Until base learning limit is reached or desired accuracy is achieved, keep repeating Step 2.   
\end{itemize}   

\begin{figure}[!h]
	\centering
	\includegraphics[width=2.5in, keepaspectratio]{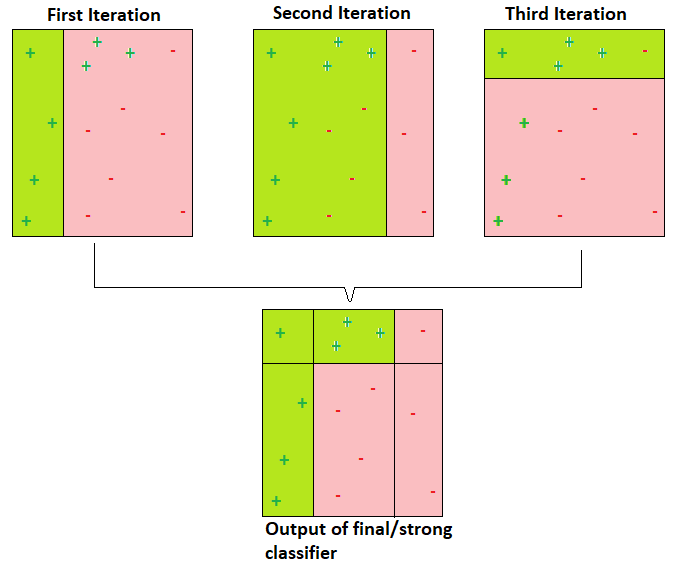}
	\caption{Working of AdaBoost Classifier}
	\label{fig_18}
\end{figure}

Ada-boost is usually used along with decision trees. Ada-boost model is created successively one after another, weight of every training instance (feature) is updated which affects the overall learning performed by next tree in the line.  After the generation of first tree, for each training instance (feature), the performance of tree is weighted i.e. how much consideration the next incoming instance (feature) will get from the new generated tree up-next. After creating all the trees, prediction is made for new data.\ignore{The performance of each tree is measured by how accurate the tree was on training data. The only disadvantage of this algorithm is that we need to remove outliers from the training data because the algorithm pays more attention to correction of the mistakes.} Fig. \ref{fig_18} shows classification done through Ada-boost technique.

\subsection{Hidden Markov Model}
Hidden Markov Model (HMM) is statistical version of Markov model and is assumed to be a Markov method having unobserved (hidden) states. In simple Markov model, observer can see the states directly and that is why Markov model have only parameters related to state transitional probabilities. On the other hand, in HMM the transitional states are not directly visible but output, which is depended on the states, is visible. As, every state has a specific probability distribution over possible output tokens, the sequence of the tokens produced by HMM gives information about the state arrangements. This phenomenon is known as pattern theory. The word hidden do not refers towards the parameters of the model. In fact, it refers towards the sequence of the states through which model passes. Parameters of the Hidden Markov Model are exactly known. HMM framework contains following components \cite{hmm}:
\begin{itemize}
		\item States, e.g. labels. Usually denoted by $T = t_{1},t_{2},...,t_{N}$.
		\item Observations, e.g. words. Usually denoted by $W = w_{1},w_{2},...,w_{N}$.  
		\item Two Special States: $t_{start}$ and $t_{end}$. These states are not directly associated with observations.
\end{itemize}

\begin{figure}[!h]
	\centering
	\includegraphics[width=3in, keepaspectratio]{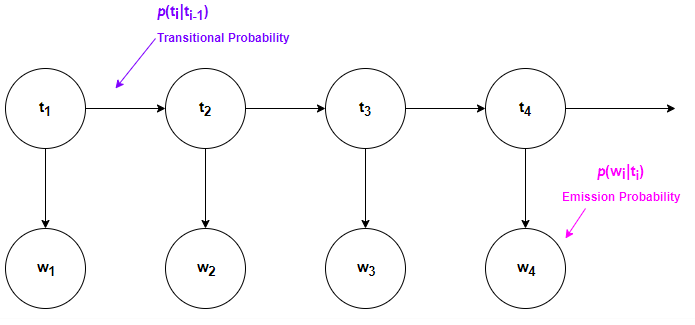}
	\caption{Transition and Emission probabilities in Hidden Markov Model}
	\label{fig_19}
\end{figure}

The states and observation related probabilities are an initial probability distribution over states, a final probability distribution over states, a matrix with the probabilities going from one state to another state, called as transition probability, and a matrix with the probabilities of an observation generated from a state, called as emission probability. Fig. \ref{fig_19} shows probabilities incur in HMM.

\ignore{A HMM having first-order has two assumptions. One, specific state probability depends upon the previous state. Mathematically;
	\begin{equation}
	P(t_{i}|t_{1},...,t_{i-1}) = P(t_{i}|t_{i-1})
	\end{equation}
	and second, output state probability only depends on state which generated the observation, not on any other observation or any state. One thing to note is that these assumptions are really closely related to Naive Bayes classifier, presented in Section \ref{NBC}.}

\subsection{Support Vector Machine Classifier}
\begin{figure}[!h]
	\centering
	\includegraphics[width=3in, keepaspectratio]{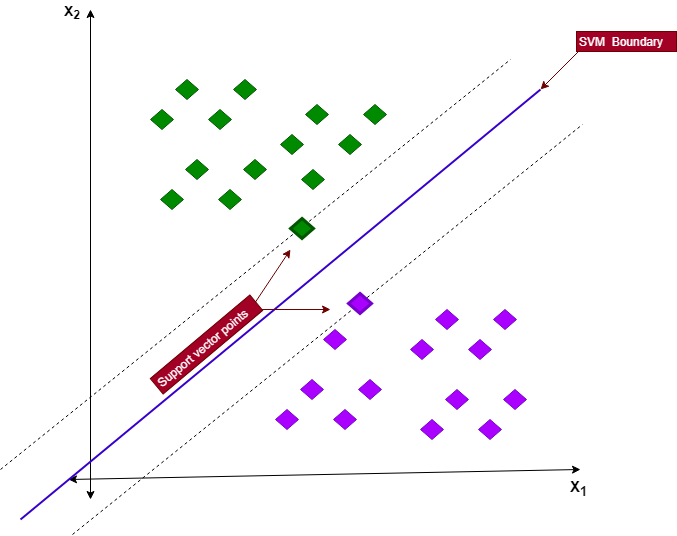}
	\caption{Maximum margin hyperplane for SVM trained model}
	\label{fig_6}
\end{figure}

Support Vector Machine (SVM) classification technique is the most precise method of solving a classification problem. These are built around a perception of margin i.e. data is separated into two classes, on each side of the hyperplane. SVM classifier is binary classifier, so for multi-class classification problem, multiple machines are trained \cite{A6}. SVM aims maximization of margin between instances (features) of the two classes and minimizing generalization error, usually incurred in other classifiers. Fig. \ref{fig_6} shows how two different set of features are classified using SVM.

Data points that lie on the margin of an optimized hyperplane are called support vector point and linear combination these point form solution to the classification problem, all other points are ignored. \ignore{Complexity of SVM does not depends upon number of instances/features present in its training set. SVM technique, even though, do tries to give maximum margin between two classes but due to presence of few misclassified features (instances) in data, SVM may not find any separating hyperplane. Solution to this problem is the use of soft margins that might accept few misclassified input instances (features). A drawback of SVM is its low training speed as we have to estimate a range of different settings and then use cross validation on the whole training dataset to find best solution.} Mathematically, SVM uses QP problem with $N$ dimensions, where $N$ shows training samples.

%%%%%%%%%%%%%%%%%%%%%%%%%%%%%%%%%%%%%%%%%%%%%%%%%%%%%%%%%%%%%%%%%%%%%%%%%%%%%%%%%%%%%%%%%%%%%%%%%%%%%%%%%%%%%%%%%%%%%%%%%%%%%%%%%%%%%%%%%%%%%%%%%%%
%%%%%%%%%%%%%%%%%%%%%%%%%%%%%%%%%%%%%%%%%%%%%%%%%%%%%%%%%%%%%%%%%%%%%%%%%%%%%%%%%%%%%%%%%%%%%%%%%%%%%%%%%%%%%%%%%%%%%%%%%%%%%%%%%%%%%%%%%%%
\section{Review of Different Machine Learning Algorithms}
\subsection{Decision Tree Classifier} \label{DT}

Schmidt et. al. \cite{R1} introduced a new, freely available, dataset for stress and its affect detection. They have recorded motion as well as physiological data using sensors attached to wrist and chest of the subjects. They examined 15 subjects in a lab. Their database is consist of three affective states i.e. neutral, stress, and amusement. Using Decision tree classifier for three class classification (neutral, stress, and amusement), they achieved the average accuracy of 53.9\% while in the case of considering Decision tree as binary classifier (stress vs non-stress), they achieved average accuracy of 74\%.

Castaldo et. al. \cite{R2}, detected mental stress by using linear as well as non-linear Heart Rate Variability (HRV) features, collected from 42 students of universities, during their oral exam (that defined stress state) and after a vacation (defined as rest state). The authors concluded that decision tree (C4.5) learning algorithm classified the two states i.e. stress and rest with an accuracy of 79\%. 

For achieving on-line assessment of user's affective states, Barreto et. al. \cite{R10}  proposed to use some of physiological signals like Blood Volume Pulse, GSR, Skin Temperature and Pupil Diameter from user to classify user's affective states by implementing some pattern learning algorithms. Authors designed a computer based "Paced Stroop Test" that acted as stimulus to any emotional stress in subject. They used Naive Bayes, Decision tree and SVM as a learning algorithms. Total of 32 subjects (students) were recruited from Florida International University and 192 feature vectors were generated (three stress; In-congruent Stroop and three non-stress; Congruent Stroop). Eleven (11) features were determined for each segmentation and only 20 samples were used as test samples, all other samples were used to train the classifiers. Accuracy achieved with Decision tree classifier when all the 11 features were used was 88.02\%. Accuracy decreased to 56.25\% when authors removed feature of Pupil Diameter and was 88.54\% when only Skin Temperature was removed from feature vector.

\subsection{Artificial Neural Network Classifier} \label{ANN}

Singh et. al. \cite{R3} presented a technique based on analysis of co-relation and proposed a mathematical function that estimates stress level of auto-mobile drivers. Stress monitoring was done using physiological parameters of the driver. The result obtained showed a strong co-relation between stress function and driver's stress level. Authors used Artificial Neural Network classifier along with some simple thresholding technique (in pre-processing stage) to classify stress states as "Low Stress", "Moderate Stress" and "High Stress" depending upon different traffic conditions. For studies, they used 10 drivers and achieved an overall accuracy of 82.5\% (combining ANN with thresholding technique).   

Cogan et. al. \cite{R4} analysed those physiological signals that can be monitor easily via wrist wearable device. They used 20 subjects for their experimentation and tried to classify four type of stress states of an individual. These types included; Physical stress, emotional stress , cognitive stress and relaxation. The authors claims to achieve 98.8\% of accurate classification of above mentioned four states by utilizing Neural Network classifier.

Verma et. al. \cite{R5} proposed a new five-layered fog cloud IoT-based student-centric monitoring of the stress to predict stress of the students in a particular context. The model computes four stressed based parameters. These parameters includes leaf node evidence, context, workload and student health trait. Authors have calculated stress indexes of 55 students and the classification technique used was Bayesian Belief Network (BNN) but for comparison they used Artificial Neural Network classifier along with some other classifiers. The accuracy they achieved via ANN classifier was 85.3\%.

\subsection{Bayesian Network Classifier} \label{BNC}

Calvo et. al. \cite{R6} used BN for stress classification and they investigated the relative effects of different parameters on the classification of mental states. These parameters included number of subjects, sampling rate, number of sessions recorded, and compared with eight different classification methods i.e. ZeroR, OneR,FT,Naive Bayes, MLP, LLR, SVM and Bayes Network. The signals choose for the study were ECG, EMG and GSR. There were only three subjects (2 Male and one Female) participated in this experiment and readings were taken in three sessions with sampling rate of 40 $Hz$ and 1$KHz$. Length of each session was 24 minutes. The accuracy achieved by Bayesian Network classifier was 81.3\% for 40$Hz$ sampling rate while for 1$KHz$ sampling rate, accuracy is not mentioned. 

\subsection{Naive Bayesian Classifier} \label{NBC}

As described in section \ref{BNC}, Calvo et. al. \cite{R6} showed the effects of variation of number of subjects, sampling rate, and number of sessions recorded on eight different classification methods. ECG, EMG and GSR signals were monitored and only three subjects (2 Male and one Female) were used in this experiment. The readings were taken in three different sessions having sampling rate of 40$Hz$ and 1$KHz$. The accuracies achieved by Naive Bayesian classifier were 66.3\% and 61.7\% for sampling rate of 40$Hz$ and 1$KHz$, respectively.

Barreto et. al. \cite{R10}, as illustrated in section \ref{DT},  proposed to utilize some of the physiological signals like Blood Volume Pulse, GSR, Skin Temperature (ST) and Pupil Diameter (PD) of the subject to classify their affective states by implementing different pattern learning algorithms. Authors designed a computer based "Paced Stroop Test" that acted as stimulus to emotional stress in the subject. They used Naive Bayes, Decision tree and SVM classifiers as a learning algorithms. Total of 32 subjects (students) were recruited from Florida International University and eleven (11) features were determined for each segmentation. Accuracy of Naive Bayes classifier with all the 11 features included was 78.65\% while it was decreased to 53.65\% when feature of Pupil Diameter was removed. The accuracy achieved, when only Skin Temperature features was removed from feature vector, was 82.81\%. 

\subsection{k-Nearest Neighbour Classifier} \label{kNN}

Mozos et. al. \cite{R8} presented a machine learning technique for automatic detection of the social stress by combining data captured using two sensors i.e. physiological and social responses. They compared the overall performance of the model using SVM, AdaBoost and kNN. Authors concluded, from experimentation, that by combining the readings form both the sensors, one can achieve an accurate classification between neutral and stressful states during Trier Social Stress Test (TSST). Eighteen subjects were used for this experimentation and accuracy achieved with kNN classifier when $k=3$ was 87.3\%. Authors also claims that their system can be used in real-time stress detection as a wearable device and allow user to determine his/her stress state at any instance.

Schmidt et. al. \cite{R1}, as described in section \ref{DT}, introduced a freely available dataset for stress and its affect detection. Authors have recorded motion as well as physiological data through different sensors attached to wrist and chest of the subjects. They investigated 15 subjects in a controlled environment. Their database was consisting of three affective states i.e. neutral, stress, and amusement. Using kNN classifier, for three class classification (neutral, stress, and amusement), authors achieved the average accuracy of 50.14\% while in the case of kNN as a binary classifier (stress vs non-stress), they achieved average accuracy of 70.5\%.

Cogan et. al. \cite{R4}, reported earlier in section \ref{ANN}, did analyses on those physiological signals which could be measured easily via wrist wearable device. Authors selected 20 subjects for their experimentation and tried to classify four type of stress states within each individual. These states were Physical stress, emotional stress , cognitive stress and relaxation. The authors claims to achieve 97.5\% of accurate classification by implementing kNN classifier.

Jebelli et. al. \cite{R9} proposed a method for automatic worker's stress detection using EEG signals. The Authors gathered EEG signals data of 11 construction field workers and preprocessed it to get high quality signals. Worker's salivary cortisol which is a stress hormone was also collected during their work and low or high stress levels were labelled. Frequency and time domain features were calculated by sliding and fixed window approach. Authors applied kNN and several SVM techniques to recognise worker's stress during their work. They achieved accuracy of 65.8\% with fixed window kNN classifier while accuracy of 61.12\% with sliding window kNN classifier. 

\subsection{Nearest Cluster Classifier} \label{NCC}

Setz et. al. \cite{R12} presented NCC as a stress classifier. Authors analysed the power of electrodermal activity (EDA) to distinguish stress due to cognitive load in work environment. Thirty three subjects were undergone a laboratory experimentation which included light cognitive load along with two stress factors related to workplace. Mental stress was induced by solving complex arithmetic problems under pressure of time, while psychosocial stress was induced by evaluating social threats. Experimentation was done using a wearable device that monitored EDA as measurement of stress of each individual. NCC and LDA classifiers were used to classify three phases (MIST, recovery and MIST+recovery phase) of feature combinations. The average accuracy of NCC classifier, when features were related to each other, was 73\% while the achieved accuracy of NCC increased to 76.5\%, when features were non-relative.     

\subsection{Learning Vector Quantization Classifier}

de Vries et. al. \cite{R13} performed a large scale study on mental stress detection in an early stages. Authors measured skin conductance, electrocardiogram and respirations rate of the subjects in a semi-controlled environment. Total of 61 subjects (20 male,41 female) participated in the experiment but only 50 participants successfully completed the training, without any technical issues. These participants were divided into three groups namely; $A (Alpha Training)$ having 18 participants, $B (random Beta Training)$ having 12 participants and $C (control, Music only)$ having 20 participants. Authors obtained an accuracy of 86.8\% by deploying Learning Vector Quantization (LVQ) method to classify stressed state from relaxation.   

\subsection{Kohonen Self-Organizing Map Classifier}

For stress classification, Ranjan et. al. \cite{R16} presented a comprehensive study on extraction and signal processing methodologies used on features of physiological signals. Authors adopted a Self Organising Map (SOM) approach to classify data in topographically distinct classes of low, medium and high stress states. For this study, 20 drivers were selected having vast experience of driving. The data collected was distributed into three stress states i.e. low, medium and high stress state. Authors analysed signals obtained form GSR, PPG (Respiration rate), ECG and HRV. For classification, KSOM classifier was used and classification was done with the accuracy of 81.6\%.

\subsection{Principal Component Analysis}

For stress monitoring, Mujeeb et. al.\cite{R17} assessed the ability of deep breathing for relaxation using wearable stress monitor. They developed a relationship between various mental stress activities and regular deep breathing patterns. Then, they used heart rate monitor, respiration rate monitor and electrodermal activity sensors to get those parameters which are consistent in ruling sympathetic nervous system. Total of 25 subjects (15 male, 10 female) participated in this study. For visualization of the data, Principle Component Analysis (PCA) was used and dimensions of the feature vector was reduced from six to two. After the whole data analysis, authors found that only 60\% of the total subjects indicated increase in the stress level caused by improper breathing and thus, concluded that it was necessary to people training about how they can improve their breathing (deal with stress).  

\subsection{Linear Discriminant Analysis} \label{LDA}

Schmidt et. al. \cite{R1}, as described in section \ref{DT}, introduced a dataset for stress and its affect detection. Authors have recorded physiological and motion related data through different sensors placed on wrist and chest of the subjects. For experiment, 15 subjects were recruited and experimentation was in a controlled environment (lab). The proposed database was consisting of three affective states i.e. neutral state, stressed state, and amusement state. Using LDA as a multi-class classifier (neutral, stress, and amusement), authors achieved the average accuracy of 62.13\% while in the case of LDA as a binary classifier (stress vs non-stress), they achieved average accuracy of 78.05\%.

Dobbins et. al. in \cite{R11} explored some alternative procedure for negative emotions detection that can provide better fidelity from psychophysiological data (such as heart rate) to dynamically labelled data derived from task of driving (like speed or road type). Existing systems mostly relays on labels obtained from individual self-report. Authors used machine learning techniques to generate labels, in two steps. First, they derived labels from individual self-reports and then labelled the data using psychophysiological activities (like Heart Rate (HR) and Pulse Transit Time (PTT), etc.) for generating dynamic labels of low vs high stress for every participant. Labelling techniques were evaluated using Linear Discriminant Analysis (LDA) and Support Vector Machine (SVM). Authors did two types of studies. In study 1, there were thirteen participants (seven females and six males) while study 2 included only eight participants (six females and two males). ECG, ear PPG and Accelerometer were used to collect data from the drivers. Total of 525,697,711 instances of data was collected for both the studies. The average accuracy (in terms of Area under the curve) of study 1 while using LDA classifier was 57.4\%. Average accuracy (in terms of Area under the curve) of study 2 was 64.2\%.

Setz et. al. \cite{R12}, as explained in section \ref{NCC}, inspected the power of electrodermal activity (EDA) for discrimination of stress state form neural state caused by cognitive load in work environment. Total of Thirty three subjects underwent a laboratory experiment. This experiment, included lenient cognitive load and two stress factors related to workplace. Mental stress was prompt by solving complex arithmetic problems under specified time. Psychosocial stress was induced by assessing social threats. For experimentation, a wearable device that monitored EDA as a parameter of stress was worn by each individual subject. NCC and LDA classifiers were used to classify three phases (MIST, recovery and MIST+recovery phase) of feature mesh. The achieved average accuracy of LDA classifier with related features was 76.5\% while when features were non-relative the accuracy of LDA classifier jumped to 80.2\%. 

\subsection{Logistic Regression}

Subhani et. al. \cite{R22} proposed a Machine learning (ML) architecture based on analysis of electroencephalogram (EEG) signals of a stressed subject. The ML framework included EEG features, ROC curve, the Bhattacharya distance, t-test, classification algorithms (Navie Bayes, SVM and Logistic Regression combined) and 10-fold cross validation. Forty two subjects, who were not taking any kind of medication that increases cardiac activities, were selected for this study. Authors achieved accuracy of 94.6\% for binary identification of stress while 83.4\% for multi-level identification.

\subsection{ZeroR and OneR classifier}

As described in section \ref{BNC} and \ref{NBC}, the effects of variation of number of subjects, sampling rate, and number of sessions recorded were examined on eight different classification methods in Calvo et. al. \cite{R6}. Signals of ECG, EMG and GSR were monitored and three subjects (2 Male and one Female) were investigated in this experiment. The readings were taken in three separate sessions with sampling rate of 40$Hz$ and 1$KHz$. The accuracies achieved by ZeroR classifier were 12.5\% for both 40$Hz$ and 1$KHz$ sampling rate, while accuracies of OneR classifier were 50.4\% and 48.3\% for 40$Hz$ and 1$KHz$, respectively.    

\subsection{Multi-Layer Perceptron Classifier}

As described in section \ref{BNC}, Calvo et. al. \cite{R6} investigated the effects of change in number of subjects, sampling rate, and number of sessions recorded on eight discreet classifiers. ECG, EMG and GSR signals were observed from only three subjects (2 Male and one Female). The readings were recorded in three different sessions with sampling rate of 40$Hz$ and 1$KHz$. The accuracies achieved by MLP classifier were 92.9\% and 97.5\% for 40$Hz$ and 1$KHz$ sampling rate, respectively.

Assessment of mental fitness while doing a task might be helpful to avoid unnecessary risks. Mohino-Herranz et. al. in \cite{R7} gathered information about mental stress by using two physiological signals i.e. ECG and electrical bio-impedance (TEB) signals. Total of 40 subjects were used to create the dataset. For classification, MLP classifier is used. Authors had designed three scenarios for the measurement. First scenario was activity identification; the proposed model distinguished different kind of activities (neutral, emotional, mental and physical) with the probability error of 21.23\%. Second scenario was emotional state; the proposed model distinguished different kind of emotional states like neutral, disgust and sadness. The probability error incurred was 4.8\% while third scenario was mental activity; the proposed model distinguished between low mental load and mental overload with the probability error of 32.3\%. 

\subsection{Genetic Algorithm} \label{GA}

Mokhayeri et. al. \cite{R14} proposed a new stress detection method that uses GA and fuzzy SVM techniques to classify stress. They analysed Pupil diameter (PD), ECG and PPG signals using soft-computation and extracted most relevant features. These features were then optimized using Gentic Algorithm (GA) and passed to Fuzzy SVM for classification between stress and relaxed states of the subject. Dataset was composed of data collected from sixty healthy peoples (25 female and 35 male). Genetic Algorithm (GA) was used to construct a template of selected eye region using average image of 100 images.

\subsection{Decision Forest} \label{DF}

Akbulut et. al. \cite{R15} designed a smart wearable device that provided continuous medical monitoring and could also create a profile showing risk of disease over time. The authors named this device as Cardiovascular Disease Monitoring (CVDiMo) device. There were two groups of 30 participants in this experiment. Different systematic tests were performed on six bio-signals collected through six different sensors i.e. ECG, Body Temperature, Pulase Oximeter, GSR, Blood Pressure and Glucometer. Beside above bio-signals, physical activity and stress levels deduced from emotional state analysis was also done to increase the performance of risk estimation. By using decision forest as a multi-class classifier, the authors achieved 96\% of accuracy. 

\subsection{Decision Jungle}

Akbulut et. al. \cite{R15}, as described in \ref{DF}, proposed a design of a smart wearable device which provided continuous medical monitoring and also creates a profile displaying disease risk over the time. Authors called this device a Cardiovascular Disease Monitoring (CVDiMo) device. Participants were distributed into two groups and total 30 people participated in this experiment. Several systematic tests were run on six bio-signals gathered form six different sensors, namely; ECG, Body Temperature, Pulse Oximeter, GSR, Blood Pressure and Gluco-meter. Along with these bio-signals, physical activity and stress levels infer from emotional state analysis were also included to increase the performance of estimation of the risk. By using decision jungle as a multi-class classifier, the overall accuracy achieved was 66\%.

\subsection{Random Forest}

Schmidt et. al. \cite{R1}, as described in section \ref{DT}, introduced a new and freely available dataset for stress detection. Authors recorded data samples of the test subjects in motion as well as their physiological data values using wearable sensors device attached to wrist and chest part. Total of 15 subjects participated in this experiment and was performed in a lab, in a controlled environment. The database was consist of three affective states, which were neutral, stress, and amusement state. Through Random Forest classifier, for three class classification (neutral, stress, and amusement), the authors achieved average accuracy of 58.24\% while in the case of binary classification (stress vs non-stress), Random Forest gave average accuracy of 77.21\%. 

Oti et. al. \cite{R20} proposed an algorithm that estimates stress level of pregnant women based on Heart Rate Variation (HRV) and simple heart rate signals. Authors distributed their algorithm in edge-enabled Internet of Things (IoT) system. They used different supervised and unsupervised learning algorithms through an unlabelled dataset gathered after 7 month monitoring. For monitoring, wearable smart wristbands were given to twenty pregnant women. Authors achieved an accuracy of 97.9\% by cross validating the proposed algorithm using Random Forest classifier. 

\subsection{One vs All Multiclass Model}

Akbulut et. al. \cite{R15}, as mentioned in \ref{DF}, demonstrated a smart wearable device that provided continuous medical health monitoring and could also create a health chart showing risk of disease over time, called as Cardiovascular Disease Monitoring (CVDiMo) device. Total of 30 participants, who were than divided into two groups, were selected for experimentation. Different routines of test were applied on six discrete bio-signals received through six different sensors including ECG, Body Temperature, Pulse Oximeter, GSR, Blood Pressure and Glucometer. Beside above bio-signals, physical activity and stress levels deduced from emotional state analysis was also taken into account to increase the performance of risk estimation. By using One vs All Multi-class Model, the authors achieved average accuracy of 72\%.

\subsection{Ada-boost}

Mozos et. al. \cite{R8}, as described in section \ref{kNN}, presented a machine learning model for automatic detection of social stress through combining data captured from physiological and social responses of the subject. They measured the performance of their model in terms of accuracies achieved from SVM, AdaBoost and kNN classifiers. Authors concluded that combining the readings of both the responses gives an accurate classification of neutral and stressful states, by conducting Trier Social Stress Test (TSST). Eighteen subjects were used for this experimentation and accuracy achieved with AdaBoost classifier was 94.3\%. Authors also claimed that their system can detect stress in real-time and is a wearable device that allows user to determine his/her stress state at any instance.

Schmidt et. al. \cite{R1}, as illustrated in section \ref{DT}, introduced a new dataset for stress. They recorded data of the subjects physiological and in motion data using sensors attached to wrist and chest part. They experimented on 15 subjects in a lab. Their database is consist of three affective states i.e. neutral, stress, and amusement. AdaBoost classifier as three class classifier i.e. neutral, stress, and amusement state, achieved average accuracy of 60.43\% while, in the case of, AdaBoost as binary classifier (stress vs non-stress), they achieved average accuracy of 77.2\%.
  
\subsection{Hidden Markov Model}

Dinges et. al. \cite{R21} developed an approach that tracks changes in facial expressions of human during a long distance space-flight. Authors applied OCR algorithms to detect low and high grade stress through facial changes during critical operations. Social and workload feedback were taken from 60 healthy peoples (29 men, 31 women), to create aa database. Stress states were tracked using salivary cortisol, heart rate and self-report ratings. Authors used dataset of face videos of the subjects during different conditions for training an OCR algorithm. For classification, Hidden Markov Model was used and accuracy of classification of high grade stress from low stress state was 75 to 88\%.  

\subsection{Support Vector Machine Classifier}

Verma et. al. \cite{R5}, as described in section \ref{ANN}, presented a five-layered fog cloud IoT-based student stress monitoring system to predict stress state of the student in a certain context. The model calculated four parameters of stress, which were leaf node evidence, context, workload, and student health trait. Authors have calculated stress indexes of 55 students. Main classification technique used in the paper was Bayesian Belief Network (BNN) but authors used SVM classifier along with some other classifiers to calculate overall performance of the algorithm. The accuracy achieved through SVM classifier was 90.8\%.

Barreto et. al. \cite{R10}, as explained in section \ref{DT},  proposed the use of some physiological signals for example Blood Volume Pulse (BVP), GSR, Skin Temperature (ST) and Pupil Diameter (PD) from subjects to classify subject's affective states by implementing few pattern learning algorithms. Authors designed a computer based test, called "Paced Stroop Test", that acted as stimulus to any emotional stress in subject. Naive Bayes, Decision tree and SVM were used as a learning algorithms. Total of 32 subjects (which were students) were recruited from Florida International University and feature vectors of 192 features (samples) were generated (three stress; In-congruent Stroop and three non-stress; Congruent Stroop). Eleven (11) features were determined for each segmentation method. Only 20 samples were used as test samples while all other samples were used to train the classifiers. Accuracy achieved with SVM classifier when all the 11 features were used was 90.10\%. Accuracy decreased to 58.85\% when authors removed feature of Pupil Diameter and got 90.1\% inputs, correctly predicted, when only Skin Temperature was removed from feature vector.

Calvo et. al. \cite{R6}, as described in section \ref{BNC}, investigated the relative effects of different specifications of classification of mental states. These specifications included number of sessions recorded, number of subjects, sampling rate, and compared with eight different classification methods which are ZeroR, OneR,FT, Naive Bayes, Bayes Net, MLP, LLR and SVM. The signals chosen for this study were ECG, EMG and GSR. There were only three subjects (2 Male and one Female) who participated in the experiment. The readings were taken in three different sessions having sampling rate of 40$Hz$ and 1$KHz$. Length of each session was not more than 24 minutes. The accuracies achieved by using SVM classifier was 94.6\% for sampling rate of 40$Hz$ and for 1$KHz$ sampling rate, accuracy was 95.8\%.

Jebelli et. al. \cite{R9}, as explained in section \ref{kNN}, presented a method for automatic detection of worker's stress using EEG signals. Authors collected EEG signals data of 11 workers, working in construction field and pre-processed it to get high quality signals. Worker's salivary cortisol, which is a stress hormone, was also collected during their work and low or high stress levels were tagged (labelled). Frequency and time domain features were intended by sliding as well as fixed window approach. Authors applied kNN and verity of SVM techniques to recognise worker's stress during their work. They achieved average accuracy of 75.9\% with fixed window SVM classifier while accuracy of 67.54\% with sliding window SVM.

Dobbins et. al. in \cite{R11}, as illustrated in section \ref{LDA}, explored some alternative procedures for negative emotions detection which can provide better fidelity using psycho-physiological data to tag data dynamically, derived from task of driving like speed or road type. Existing systems relays on tags acquired from individual self-report but authors used machine learning techniques to generate tags. Tags generation was done in two steps. First, they derived tags from individual self-reports and then they tagged the data using psycho-physiological activities (like Heart Rate (HR) and Pulse Transit Time (PTT), etc.) for generating dynamic tags of low vs high stress for every participant. Tagging techniques were evaluated using Linear Discriminant Analysis (LDA) and Support Vector Machine (SVM). Authors did two types of studies. In study 1, there were thirteen participants (seven females and six males) while study 2 included only eight participants (six females and two males). ECG, ear PPG and Accelerometer were used to gather data from drivers. Total of 525,697,711 instances of data was gathered for both studies. The average accuracy (in terms of Area under the curve) of study 1 while using SVM classifier was 58.8\%. Average accuracy (in terms of Area under the curve) of study 2 was 65.8\%.

Mokhayeri et. al. \cite{R14}, as illustrated in section \ref{GA}, demonstrated a novel procedure for stress detection. They utilized Pupil diameter (PD), ECG and PPG signals using soft-computation and extracted most relevant features. These features were then optimized using Gentic Algorithm (GA)and passed to Fuzzy SVM for classification of stress and relaxed states of the subject. Data was collected from sixty healthy subjects (25 female and 35 male). The detection of eye region was done by constructing a template by using average image of 100 images. After the segmentation (done by GA algorithm), SVM classifier was applied to classify between stressed and relaxation states. The achieved accuracy of SVM classifier was 90.1\%.

Salafi et. al. \cite{R18} presented a study on an unobtrusive wearable stress monitoring device which had three physiological sensors namely; heart rate monitoring sensor (ECG, PPG), skin conductance (GSR) and skin temperature monitoring sensor. Total of 50 participants (23 male, 27 female) form National University of Singapore were selected to gather training data. Support Vector Machine (SVM) learning algorithm was applied to the data and accuracy of 91.26\% was achieved.

Betti et. al. \cite{R19} developed and tested the ability of a wearable sensor devices to detect human stress using ECG, EEG and EDA signals. The authors also assessed co-relation of detected stress signals and salivary cortisol level, which is considered as a reliable bio-marker for stress. For experiment, 15 healthy subjects wore a set of three commercially available sensors to record readings of physiological signals when a Maastricht Acute Stress Test (MAST) was conducted. Salivary samples were also collected, at random, during the test. Statistical analysis were performed using Support Vector Machine (SVM) classifier. Classification algorithm, provided an accuracy of 86\% on selected (significant) features.
\begin{table*}
	\centering
	\caption{Comparison Table of Machine Learning Approach for Stress Monitoring and Identification}
	\begin{tabular}{|c|c|c|c|c|c|}
		\hline
		\textbf{S.No} & \textbf{Method/Technique} & \textbf{Ref} & \textbf{No. of subjects} & \textbf{Signals/Devices Used} & \textbf{Avg\_Accuracy (\%)} \\ \hline \rule{0pt}{10pt}
			1 & \multirow{2}{*}{AdaBoost} & \cite{R1} & 15 & Motion as well as physiological data using sensors & 77.2 \\ \cline{1-1} \cline{3-6} \rule{0pt}{10pt}
			2 &  & \cite{R8} & 18 & Physiological and social responses & 94.3 \\ \hline \rule{0pt}{10pt}
			3 & \multirow{3}{*}{ANN} & \cite{R3} & 10 & ECG and PPG & 82.5 \\ \cline{1-1} \cline{3-6} \rule{0pt}{10pt}
			4 &  & \cite{R4} & 20 & ECG and PPG & 98.8 \\ \cline{1-1} \cline{3-6} \rule{0pt}{10pt}
			5 &  & \cite{R5} & 55 & Leaf node evidence, context, workload, and student health trait. & 85.3 \\ \hline \rule{0pt}{15pt}
			6 & \multirow{2}{*}{Decision Forest} & \cite{R15} & 30 & \begin{tabular}[c]{@{}c@{}}ECG, Body Temperature, Pulse Oximeter, \\ GSR, Blood Pressure, and Glucometer\end{tabular} & 96 \\ \cline{1-1} \cline{3-6} \rule{0pt}{15pt}
			7 &  & \cite{R15} & 30 & \begin{tabular}[c]{@{}c@{}}ECG, Body Temperature, Pulse Oximeter, \\ GSR, Blood Pressure, and Glucometer\end{tabular} & 66 \\ \hline \rule{0pt}{10pt}
			8 & \multirow{3}{*}{Decision Tree} & \cite{R1} & 15 & Motion as well as physiological data using sensors & 74 \\ \cline{1-1} \cline{3-6} \rule{0pt}{10pt}
			9 &  & \cite{R10} & 34 & Blood Volume Pulse, GSR, Skin Temperature and Pupil Diameter & 88.02 \\ \cline{1-1} \cline{3-6} \rule{0pt}{10pt}
			10 &  & \cite{R2} & 42 & Heart Rate Variability (HRV) & 79 \\ \hline \rule{0pt}{10pt}
			11 & Genetic  Algorithm & \cite{R14} & 60 & Pupil diameter (PD), ECG and PPG signals & N/A \\ \hline \rule{0pt}{10pt}
			12 & HMM & \cite{R21} & 60 & \begin{tabular}[c]{@{}c@{}}Salivary cortisol, heart rate, and self-report ratings + \\ moment of eyebrows and mouth\end{tabular} & 75-88 \\ \hline \rule{0pt}{10pt}
			13 & \multirow{4}{*}{kNN} & \cite{R1} & 15 & Motion as well as physiological data using sensors & 70.5 \\ \cline{1-1} \cline{3-6} \rule{0pt}{10pt}
			14 &  & \cite{R4} & 20 & ECG and PPG & 97.5 \\ \cline{1-1} \cline{3-6} \rule{0pt}{10pt}
			15 &  & \cite{R8} & 18 & Physiological and social responses & 87.3 \\ \cline{1-1} \cline{3-6} \rule{0pt}{10pt}
			16 &  & \cite{R9} & 11 & EEG signals + salivary cortisol & 63.46 \\ \hline \rule{0pt}{10pt}
			17 & KSOM & \cite{R16} & 20 & GSR, PPG (Respiration rate), ECG and HRV. & 81.6 \\ \hline \rule{0pt}{10pt}
			18 & LVQ & \cite{R13} & 50 & Skin conductance, electrocardiogram, and respiration rate & 86.8 \\ \hline \rule{0pt}{15pt}
			19 & \multirow{3}{*}{LDA} & \cite{R1} & 15 & Motion as well as physiological data using sensors & 78.05 \\ \cline{1-1} \cline{3-6} \rule{0pt}{10pt}
			20 &  & \cite{R11} & 21 & Heart Rate (HR) and Pulse Transit Time (PTT) & \begin{tabular}[c]{@{}c@{}}60.8   approx\end{tabular} \\ \cline{1-1} \cline{3-6} \rule{0pt}{10pt}
			21 &  & \cite{R12} & 33 & Electrodermal activity (EDA) & 78.35 approx \\ \hline \rule{0pt}{15pt}
			22 & Logistic Regression & \cite{R22} & 42 & Electroencephalogram (EEG) signals & 89 approx \\ \hline \rule{0pt}{10pt}
			23 & \multirow{2}{*}{MLP} & \cite{R7} & 40 & ECG and electrical bio-impedance (TEB) signals & 67.7 \\ \cline{1-1} \cline{3-6} \rule{0pt}{10pt}
			24 &  & \cite{R6} & 3 & ECG, EMG, and GSR & 95.2 approx \\ \hline \rule{0pt}{10pt}
			25 & \multirow{2}{*}{Naïve Bayes} & \cite{R10} & 33 & Blood Volume Pulse, GSR, Skin Temperature and Pupil Diameter & 78.65 \\ \cline{1-1} \cline{3-6} \rule{0pt}{10pt}
			26 &  & \cite{R6} & 3 & ECG, EMG, and GSR & 64 approx \\ \hline \rule{0pt}{15pt}
			27 & \begin{tabular}[c]{@{}c@{}}Nearest Class Center (NCC)\end{tabular} & \cite{R12} & 33 & Electrodermal activity (EDA) & \begin{tabular}[c]{@{}c@{}}74.52\\   approx\end{tabular} \\ \hline  \rule{0pt}{10pt}
			28 & OneR & \cite{R6} & 3 & ECG, EMG, and GSR & 49.35 approx \\ \hline \rule{0pt}{10pt} 
			29 & One-v-all & \cite{R15} & 30 & \begin{tabular}[c]{@{}c@{}}ECG, Body Temperature, Pulse Oximeter, GSR, Blood \\ Pressure and Glucometer\end{tabular} & 72 \\ \hline \rule{0pt}{10pt}
			30 & PCA & \cite{R17} & 25 & \begin{tabular}[c]{@{}c@{}}Heart rate monitor, respiration rate monitor and electrodermal \\ activity sensors\end{tabular} & 60 \\ \hline \rule{0pt}{15pt}
			31 & \multirow{2}{*}{Random Forest} & \cite{R1} & 15 & Motion as well as physiological data using sensors & 77.21 \\ \cline{1-1} \cline{3-6} \rule{0pt}{10pt}
			32 &  & \cite{R20} & 20 & Heart Rate Variation (HRV) and simple heart rate signals. & 97.2 \\ \hline \rule{0pt}{10pt}
			33 & \multirow{8}{*}{SVM} & \cite{R10} & 32 & Blood Volume Pulse, GSR, Skin Temperature and Pupil Diameter & 90.1 \\ \cline{1-1} \cline{3-6} \rule{0pt}{10pt}
			34 &  & \cite{R11} & 21 & \begin{tabular}[c]{@{}c@{}}Heart Rate (HR) and Pulse Transit Time (PTT)\end{tabular} & \begin{tabular}[c]{@{}c@{}}62.3  approx\end{tabular} \\ \cline{1-1} \cline{3-6} \rule{0pt}{10pt}
			35 &  & \cite{R14} & 60 & Pupil diameter (PD), ECG and PPG signals & 90.1 \\ \cline{1-1} \cline{3-6} \rule{0pt}{10pt}
			36 &  & \cite{R18} & 50 & \begin{tabular}[c]{@{}c@{}}Heart rate monitoring sensor (ECG, PPG), skin conductance (GSR) and \\ skin temperature monitoring sensor\end{tabular} & 91.26 \\ \cline{1-1} \cline{3-6} \rule{0pt}{10pt}
			37 &  & \cite{R19} & 15 & ECG, EEG and EDA signals + Salivary samples & 86 \\ \cline{1-1} \cline{3-6} \rule{0pt}{10pt}
			38 &  & \cite{R5} & 55 & Leaf node evidence, context, workload, and student health trait. & 90.8 \\ \cline{1-1} \cline{3-6} \rule{0pt}{10pt}
			39 &  & \cite{R6} & 3 & ECG, EMG, and GSR & 95.2 approx \\ \cline{1-1} \cline{3-6} \rule{0pt}{10pt}
			40 &  & \cite{R9} & 11 & EEG signals + salivary cortisol & 71.72 \\ \hline \rule{0pt}{10pt}
			41 & ZeroR & \cite{R6} & 3 & ECG, EMG, and GSR & 12.5 approx \\ \hline 
		\end{tabular}
		\begin{tabular}{l}
		Note: approx means it is average of all the accuracies achieved by author/s in different scenarios. OneR and ZeroR have accuracies less than 50\% \\ (selection criteria) but are baseline algorithms that is why they are included here.
	\end{tabular}
\end{table*}

% Note that IEEE does not put floats in the very first column - or typically
% anywhere on the first page for that matter. Also, in-text middle ("here")
% positioning is not used. Most IEEE journals use top floats exclusively.
% Note that, LaTeX2e, unlike IEEE journals, places footnotes above bottom
% floats. This can be corrected via the \fnbelowfloat command of the
% stfloats package.

%\section{Result and Discussion}
\section{Discussion and Conclusion}
In mental health research, psychological and sociological stress are considered as one of the most important as well as most complex areas of the current century. High stress does becomes thread to health of a person. The reasons behind this extensive stress are complex personal, social and ecological environment, diversity in expressing stress as well as multiple transactions of human due to his/her surroundings. Even-though stress is a routine characteristic of  life, nowadays, but if it becomes continuous and increasing, individual might show problematic symptoms which threatens their health as well as peoples in their surroundings.    

This study focused on classification of stress states. Total 21 different machine learning algorithms are discussed in this review paper that uses different parameters for training and prediction of stress, see Table I. All the parameters considered, are correlated with the stress. Few of them are distinct parameters, for example GSR and HR, while some are in conjunction with other parameters to monitor and recognize stress, for example Skin temperature (ST) and EMG. But only using parameters shown in Table I, stress can not be defined. We also require information about the context to interpret the data collected from the sensors and also to understand what was going on at the time of reading collection. This context information can be gathered using mobile phones (IoT based) or from computers as we devote a lot of time using them for our daily life work.

In this study we gave a better in look of state-of-the-art machine learning algorithms and their use as stress level classifier. Table I show 21 algorithms tested on the task of stress level monitoring. We can see that only 6 out of 21, namely; SVM, Random Forest, kNN, MLP, Ada-Boost and Decision Forest machine learning algorithm were able to recognize and classify stress state of a subject with the accuracy of more than 90\%. In terms of consistent higher classification rate, SVM classifier can be considered as best classifier for stress monitoring procedures but one have to see number of subjects and type of sensors (signals) required for training and testing of the classifier. The question that which classification algorithm is best for utilization in stress monitoring device remains in place as different people have different perception about the usage of machine learning algorithms. For some people, simpler algorithms are the best while some people thinks no matter how complex an algorithm is but the accuracy should be higher. For us, it is a kind of trade-off between computation time vs accuracy vs price of the device. 

% Please add the following required packages to your document preamble:
% \usepackage{booktabs}
% Please add the following required packages to your document preamble:
% \usepackage{booktabs}

% if have a single appendix:
%\appendix[Proof of the Zonklar Equations]
% or
%\appendix  % for no appendix heading
% do not use \section anymore after \appendix, only \section*
% is possibly needed

% use appendices with more than one appendix
% then use \section to start each appendix
% you must declare a \section before using any
% \subsection or using \label (\appendices by itself
% starts a section numbered zero.)
%

%\appendices
%\section{Proof of the First Zonklar Equation}
%Appendix one text goes here.
%
%% you can choose not to have a title for an appendix
%% if you want by leaving the argument blank
%\section{}
%Appendix two text goes here.
%
%
%% use section* for acknowledgement
%\section*{Acknowledgment}
%
%
%The authors would like to thank...

% Can use something like this to put references on a page
% by themselves when using endfloat and the captionsoff option.
\ifCLASSOPTIONcaptionsoff
  \newpage
\fi

\bibliographystyle{ieeetr}

% that's all folks
\end{document}